\documentclass[review]{elsarticle}

\usepackage{xcolor}

\usepackage{url}
\usepackage{latexsym}
\usepackage[noend]{algpseudocode}
\usepackage{tabularx}
\usepackage{multirow}
\usepackage{amsmath}
\usepackage[normalem]{ulem}
\usepackage{algorithm}
\usepackage{algorithmicx}
\usepackage{multicol}
\usepackage{appendix}
\usepackage{amssymb}
\usepackage{amsthm}
\usepackage{epsfig}

\newtheorem{theorem}{Theorem}

\journal{Information Sciences}

\begin{document}

\begin{frontmatter}

\title{Towards Easier and Faster Sequence Labeling for Natural Language Processing:\\ A Search-based Probabilistic Online Learning Framework (SAPO)}



\author[eecs,moe,bigdata]{Xu Sun\corref{mycorrespondingauthor}}
\cortext[mycorrespondingauthor]{Corresponding author}
\ead{xusun@pku.edu.cn}

\author[eecs,moe]{Shuming Ma}
\ead{shumingma@pku.edu.cn}

\author[eecs,moe]{Yi Zhang}
\ead{zhangyi16@pku.edu.cn}

\author[eecs,moe]{Xuancheng Ren}
\ead{renxc@pku.edu.cn}

\address[eecs]{School of Electronics Engineering and Computer Science, Peking University}
\address[moe]{MOE Key Laboratory of Computational Linguistics, Peking University}
\address[bigdata]{Center for Data Science, Peking University}

\begin{abstract}
There are two major approaches for sequence labeling. One is the probabilistic gradient-based methods such as conditional random fields (CRF) and neural networks (e.g., RNN), which have high accuracy but drawbacks: slow training, and no support of search-based optimization (which is important in many cases). The other is the search-based learning methods such as structured perceptron and margin infused relaxed algorithm (MIRA), which have fast training but also drawbacks: low accuracy, no probabilistic information, and non-convergence in real-world tasks. We propose a novel and ``easy" solution, a search-based probabilistic online learning method, to address most of those issues. The method is ``easy'', because the optimization algorithm at the training stage is as simple as the decoding algorithm at the test stage. This method searches the output candidates, derives probabilities, and conducts efficient online learning. We show that this method, which is easy to implement, can support search-based optimization and obtain top accuracy with fast training and theoretical guarantee of convergence.
Experiments on well-known tasks show that our method has better accuracy than CRF and BiLSTM\footnote{The SAPO code is released at \url{https://github.com/lancopku/SAPO}}.\\
\end{abstract}


\end{frontmatter}


\section{Introduction}

Sequence labeling models are popularly used to solve structure dependent problems in a wide variety of application domains, including natural language processing, bioinformatics, speech recognition, and computer vision.
To solve those problems, many sequence labeling methods have been developed, most of which are from two major categories.
One is the probabilistic gradient-based learning methods such as conditional random fields (CRF) \cite{LaffertyMcCallum01} and neural networks (e.g., RNN) \cite{Hochreiter1997}.
The other is the search-based learning methods such as the margin infused relaxed algorithm (MIRA) \cite{jmlr/CrammerS03} and structured perceptrons \cite{Collins2002}. Other related work on sequence labeling  includes maximum margin Markov networks \cite{Taskar2003} and structured support vector machines \cite{Tsochantaridis04}.

As for the probabilistic gradient-based learning methods such as CRF and RNN, they have high accuracy because of the exact computation of the gradient and probabilistic information. Nevertheless, those methods have critical drawbacks.

First, the probabilistic gradient-based methods typically do not support search-based optimization (search-based learning or decoding-based learning), which is important in sequence labeling problems with emphasis on the learning speed (e.g., for large-scale datasets). In the tasks with complex structures, the gradient computation is usually quite complicated and sometimes even intractable. This is mainly because dynamic programming for computing gradient is hard to scale for large-scale datasets. On the other hand, the search technique is easier to scale to large-scale datasets. 
This is because search-based learning is much simpler than gradient-based learning~\cite{ShaPereira2003,VishwanathanSchraudolph06,Sun_NIPS2014} --- just search the promising output candidates and compare them with the oracle labels and update the weights accordingly.
	

Another category of sequence labeling methods is the search-based learning methods (decoding-based learning) such as structured perceptrons\footnote{The perceptron model used in the manuscript is the structured perceptron which includes features that incorporate transition features (structural information). The decoding of the structured perceptrons has to apply the Viterbi algorithm, because of the introduction of transition features. The decoding is the same with CRF models, which indeed considers the transition features. It is very different from the ordinary non-structured perceptron models, which can use greedy decoding because of the lack of transition features.} and MIRA. A major advantage of those methods is that they support search-based learning such that the gradient is not needed and the learning is done by simply searching and comparing the promising output candidates with the oracle labels, and updating the model weights accordingly. As a by-product of the avoidance of gradient computation, those methods have faster training speed compared with probabilistic gradient-based learning methods such as CRF. However, there are also severe drawbacks of the existing search-based learning methods:

\begin{itemize}
	\item First, the existing search-based learning methods such as structured perceptrons and MIRA have relatively low accuracy, compared with the probabilistic gradient-based learning methods such as CRF and RNN.
	
	\item Second, when applied to most of the real-world tasks, those search-based learning methods are non-convergent, i.e., they diverge in the training. As large margin classification models, theoretically those search-based learning methods have some convergent properties dependent on strict separability conditions. However, those strict separability conditions are not satisfiable in most real-world tasks, as demonstrated in many lines of prior work \cite{SunLWL14}. We will also show in the experiments that those search-based learning methods diverge dramatically as the training goes on, which makes the model accuracy go worse and worse.
	
	\item The existing search-based methods do not support probabilistic information. The magnitude of model weights grows dramatically as training goes on, and there is no reliable probabilistic information that can be derived. We will also show the curves of the model weight magnitude in the experiments.
\end{itemize}

To address those issues, we propose a novel and ``easy" solution, a \underline{\textbf{s}}earch-b\underline{\textbf{a}}sed \underline{\textbf{p}}robabilistic \underline{\textbf{o}}nline learning framework (SAPO), which can fix almost all of those drawbacks. The method is ``easy'' because the optimization algorithm at the training stage is as simple as the decoding algorithm at the test stage. The proposed method searches the top-$n$ output candidates, derives probabilities based on the searched candidates, and conducts fast online learning by updating the model weights.

We show that the proposed method is of fast training speed which is comparable with structured perceptrons and MIRA, able to support search-based optimization and no need to calculate gradient, very easy to implement, with top accuracy which is even better than CRF and BiLSTM, and with theoretical guarantees of convergence towards the optimum given reasonable conditions. 
Experiments on well-known tasks show that our method has better accuracy than CRF and BiLSTM and almost as fast training speed as structured perceptrons and MIRA.

The contributions of this work are as follows:

\begin{itemize}
	\item On the methodology side, we propose a general purpose search-based probabilistic online learning framework SAPO for sequence labeling. We show that SAPO can address a variety of issues of existing methods. Compared with probabilistic gradient-based learning methods such as CRF and RNN, the proposed method supports search-based learning such that avoiding complex gradient calculation, and with extra advantages on accuracy and training speed. Compared with search-based learning methods such as structured perceptron and MIRA, SAPO has much higher accuracy.
	
	\item For the proposed method, we provide theoretical and empirical justifications of convergence, even if the data is linearly non-separable. We provide a novel theoretical analysis on the convergence of the proposed method, which does not rely on the assumption of linearly separable margin. On the other hand, structured perceptron and MIRA diverge in real-world tasks, which have linearly non-separable datasets, as to be shown in our experiments.
	
	\item On the application side, for several benchmark natural language processing tasks, including part-of-speech tagging, biomedical entity recognition, and phrase chunking, our simple search-based learning method can easily beat the strong baseline systems on those competitive tasks with faster speed.
\end{itemize}

\section{Proposed method}

We first describe the proposed search-based probabilistic online learning algorithm SAPO; then, we compare SAPO with existing methods.

\subsection{Search-based probabilistic online learning}

\begin{algorithm}[tb]
	\caption{Search-based Probabilistic Online Learning Algorithm (SAPO)}\label{algo1}
	\begin{algorithmic}[1]
		
		\State {\textbf{input}: top-$n$ search parameter $n$, regularization strength $\lambda$, learning rate $\gamma$}
		\Repeat
		\State {Draw a sample $\pmb z=(\pmb x, \pmb y^*)$ at random from training set $S$}
		\State {Based on $\pmb w$, search the top-$n$ outputs $Y_n=\{\pmb y_{1}, \pmb y_{2}, \cdots, \pmb y_{n}\}$}
		\State {For every $\pmb y_{k} \in Y_n$, compute the probability $P_{k}= P(\pmb y_{k} |\pmb x, \pmb w)$}
		\State {For every $\pmb y_{k} \in Y_n$, update the weights by $\pmb w \leftarrow \pmb w - \gamma P_{k} \pmb F(\pmb x, \pmb y_{k})$}
		\State {For $\pmb y^*$, update the weights by $\pmb w \leftarrow \pmb w + \gamma \pmb F(\pmb x, \pmb y^*)$}
		\State {Regularize the weights by $\pmb w \leftarrow \pmb w - \frac {\gamma \lambda} {|S|} \nabla R(\pmb w)$}
		\Until {Convergence}
		\State \Return {the learned weights $\pmb w^*$}
		
	\end{algorithmic}
\end{algorithm}

The proposed search-based probabilistic online learning algorithm SAPO has the key schemes as follows: top-$n$ search (either exact search or approximate search), a scheme for calculating probabilities, perceptron-style update for weights, and a regularizer on weights. We introduce the technical details of the key schemes as follows, after which we summarize the SAPO algorithm in Algorithm~\ref{algo1}.

First, SAPO draws a training sample $\pmb z=(\pmb x, \pmb y^*)$ at random from training set $S$, and searches for the top-$n$ outputs:
$$
Y_n=\{\pmb y_{1}, \pmb y_{2}, \cdots, \pmb y_{n}\}
$$
In this work, each output $\pmb y$ is a structured label sequence $\pmb y = \{y_1, y_2, \cdots, y_l\}$, where $l$ is the number of labels in the sequence. There are many methods to realize top-$n$ search. One method uses the A* search algorithm \cite{HartNilsson68}.
An $A^*$ search algorithm with a
Viterbi heuristic function can be used to produce top-$n$ outputs
one-by-one in a efficient manner. We use the backward Viterbi algorithm \cite{Viterbi67} to
compute the admissible heuristic function for the forward-style
$A^*$ search.
This way, we can produce the
top-$n$ taggings efficiently.\footnote{Note that, although our search is ``exact'' top-$n$ search, ``exact'' top-$n$ search is not strictly required in the SAPO framework. In other words, we can replace exact A* search with non-exact beam search scheme for the SAPO algorithm. In the experiments we test both exact A* search and non-exact beam search with pruning (beam size is 50), and we find that there is almost no difference on the experimental results.}

Then, for every $\pmb y_{k} \in Y_n$, compute the probability in a log-linear fashion:
\begin{equation}\label{eq1}
P_{k}
\triangleq P(\pmb y_{k} |\pmb x, \pmb w)
\triangleq \frac {\exp [\pmb w^T \pmb F(\pmb x, \pmb y_{k})]} {\sum_{\forall \pmb y \in Y_n} \exp [\pmb w^T \pmb F(\pmb x, \pmb y)]}
\end{equation}
where $\pmb w$ is the vector of the model weights, $\pmb F(\pmb x, \pmb y_{k})$ is the feature vector based on $\pmb x$ and $\pmb y_k$, and $Y_n$ is simply the top-$n$ outputs defined before. With this definition, we can see that $\sum_{k=1}^n P_k = 1$. That is, we use top-$n$ search results to estimate the probability distribution, which is typically defined as:
\begin{equation}\label{eq2}
P(\pmb y_{k} |\pmb x, \pmb w)
\triangleq \frac {\exp [\pmb w^T \pmb F(\pmb x, \pmb y_{k})]} {\sum_{\forall \pmb y} \exp [\pmb w^T \pmb F(\pmb x, \pmb y)]}
\end{equation}
As we can see, the only difference is the normalizer --- we use top-$n$ search results to estimate the normalizer.
With the growth of $n$, this probability estimation in (\ref{eq1}) goes more and more accurate towards the traditional probability in (\ref{eq2}). On the theoretical side, we will show in the theoretical analysis that this probability estimation can be arbitrarily-close to the traditional probability by using a proper $n$, and the SAPO algorithm is guaranteed to converge towards the optimum weights $\pmb w^*$ with an arbitrarily-close distance, given reasonable conditions. On the empirical side, we will show in experiments that the probability estimation is good enough for most real-world tasks even with $n=5$ or $n=10$.

After that, SAPO updates the weights in a perceptron fashion. For every $\pmb y_k \in Y_n$, the weights are updated as follows:
\begin{equation}
\pmb w \leftarrow \pmb w - \gamma P_{k} \pmb F(\pmb x, \pmb y_{k})
\end{equation}
As we can see, this is similar to the perceptron style update, except with an additional learning rate $\gamma$ and a probabilistic scale $P_{k}$. On the other hand, for the oracle tagging $\pmb y^*$, the weights are updated by
\begin{equation}
\pmb w \leftarrow \pmb w + \gamma \pmb F(\pmb x, \pmb y^*)
\end{equation}
As we can see, this is also similar to the perceptron style update. There is no need to use a probability scale here, because the probability is 1.

Finally, SAPO uses a weight regularizer with regularization strength $\lambda$, just like the stochastic regularization adopted in stochastic gradient descent (SGD) \cite{BottouC03,conf/nips/RechtRWN11,SunLWL14}. Following the regularization scheme of SGD, the regularization strength turns to $\lambda/|S|$ in the online learning setting \cite{BottouC03,conf/nips/RechtRWN11,SunLWL14}. Also, the regularization should be scaled with the learning rate $\gamma$. Thus, by using a regularizer denoted as $R(\pmb w)$, the regularization step is as follows:
\begin{equation}
\pmb w \leftarrow \pmb w - \frac {\gamma \lambda} {|S|} \nabla R(\pmb w)
\end{equation}
The regularizer $R(\pmb w)$ can be $L_2$, $L_1$, or other regularization terms. For simplicity, in this work we use the most widely used $L_2$ regularizer (a Gaussian prior).

The SAPO algorithm is summarized in Algorithm~\ref{algo1}.

\subsection{Comparison and discussion}\label{sec:discuss}

Among the existing sequence labeling methods, the most similar and related methods to SAPO are structured perceptrons \cite{Collins2002} and CRF \cite{LaffertyMcCallum01}.

If we compare SAPO with the structured perceptron \cite{Collins2002} and CRF \cite{LaffertyMcCallum01} with stochastic training, it is interesting to see that SAPO is like a ``unification'' of the structured perceptron and the stochastically trained CRF.
The differences between CRF, structured perceptron, and SAPO mainly lie in how they estimate the parameters using the sequential information in the training phase. Structured perceptron updates the parameters only using the gold sequence path. CRF updates the parameters using an expectation of all the possible sequence paths. SAPO updates the parameters using the top-n possible sequence paths.
If we neglect the learning rate and the regularizer term of SAPO, the structured perceptron algorithm \cite{Collins2002} can be seen as an extreme case of SAPO with $n=1$ (i.e., using top-1 search instead of top-$n$ search). On the other hand, the stochastically trained CRF can be seen as another extreme case of SAPO with exponentially big $n$ that enumerates over all possible output taggings (the only difference is that CRF uses dynamic programming instead of top-$n$ search).

In other words, structured perceptron can be seen as SAPO with extremely small $n$, and CRF can be seen as SAPO with extremely big $n$. We argue that SAPO is more natural than both structured perceptrons and CRF --- we should use a moderate value of $n$ instead of an extremely small $n$ (structured perceptrons) or an extremely huge $n$ (CRF). As we will show in experiments and theoretical analysis, an extremely small $n$ like structured perceptron will lead to low accuracy and non-convergent training, and an extremely large $n$ like CRF will also lead to loss of accuracy (due to the overfitting of probabilities) and high computational cost.
In practice, we find it good enough to use $n=5$ or $n=10$ for real-world tasks.

The MIRA algorithm also has a variation of Nbest MIRA which also uses top-$n$ search \cite{jmlr/CrammerS03}. Interestingly, it is also good enough to use $n=5$ or $n=10$ for Nbest MIRA \cite{jmlr/CrammerS03,acl/McDonaldCP05,jmlr/Chiang12}.
Nevertheless, SAPO is substantially different compared with Nbest MIRA. The major difference is that SAPO has probability estimation of different outputs while Nbest MIRA does not. Nbest MIRA treats different outputs equally without probability difference, which is why CRF cannot be seen as a special case of Nbest MIRA. Even if Nbest-MIRA uses extremely huge $n$ in top-$n$ search, it is not equivalent to CRF, and the difference is substantial. Also, there are other differences between SAPO and Nbest MIRA. For example, SAPO has the regularizer term and the learning rate and has no need to use the ``minimum change'' optimization criterion of MIRA during weight update.

\section{Theoretical analysis}

Here we give theoretical analysis on the objective function, update term, convergence conditions, and convergence rate.

\subsection{Objective function and update term}
Here we analyze the equivalent objective function of SAPO and the update term of SAPO.
The SAPO algorithm (Algorithm \ref{algo1}) is a search-based optimization algorithm so that there is no need to compute the gradient of an objective function, and there is no explicit objective function used in the SAPO algorithm. Nevertheless, interestingly, we show that the SAPO algorithm is convergent and it converges towards the optimum weights $\pmb w^*$ which maximizes the objective function as follows:\footnote{The subscript of $\pmb y$ is overloaded here. For clarity throughout, $\pmb y$ with subscript $i$ and usually with the $*$ mark refers to the tagging of the $i$'th indexed training sample (e.g., $\pmb y_i^*$), and $\pmb y$ with subscript $k$ refers to the $k$'th output of the search (e.g., $\pmb y_k$).}
\begin{equation}\label{eq3}
\text{maximize}_{\pmb w} \sum_{i=1}^m \log P(\pmb y_i^*|\pmb x_i, \pmb w) -  \lambda R(\pmb w)
\end{equation}
where $m$ is the number of training samples, i.e., $m=|S|$, and $R(\pmb w)$ is a weight regularization term for controlling overfitting. This objective function is similar to that of CRF. Equivalently, for the convenience of convex-based analysis, we denote the objective function $f(\pmb w)$ as the negative form of (\ref{eq3}):
\begin{equation}
f(\pmb w) = - \sum_{i=1}^m \log P(\pmb y_i^*|\pmb x_i, \pmb w) + \lambda  R(\pmb w)
\end{equation}
We show that the SAPO algorithm converges towards the optimum $\pmb w^*$ which minimizes the convex objective function of $f(\pmb w)$:
\begin{equation}\label{obj}
\pmb w^* = \text{minimize}_{\pmb w} f(\pmb w)
\end{equation}
To clarify the theoretical analysis, we compare SAPO with the SGD (stochastic gradient descent) training scheme. Recall that the weight update has the following form in SGD \cite{BottouC03,conf/nips/RechtRWN11}:\footnote{In practice, SGD and SAPO can use decayed learning rate or fixed learning rate. Following \cite{conf/nips/RechtRWN11,Sun_NIPS2014}, for the convenience of theoretical analysis, our theoretical analysis is more focused on SGD and SAPO with fixed learning rate.}
\begin{equation}
\pmb w \leftarrow \pmb w - \gamma \nabla f_{\pmb z}(\pmb w)
\end{equation}
where $\nabla f_{\pmb z}(\pmb w)$ is the stochastic gradient of $f(\pmb w)$ based on the sample $\pmb z$, which has the following form when using the CRF objective function:
\begin{equation}\label{eq4}
\begin{split}
\nabla f_{\pmb z}(\pmb w)
&= - \Big \{ \pmb F(\pmb x, \pmb y^*) - \sum_{\forall \pmb y} P(\pmb y|\pmb x, \pmb w) \pmb F(\pmb x, \pmb y) - \frac \lambda {|S|}  \nabla R(\pmb w)  \Big \}\\
&= - \Big \{ \pmb F(\pmb x, \pmb y^*) - \sum_{\forall \pmb y} \frac {\exp [\pmb w^T \pmb F(\pmb x, \pmb y)]} {\sum_{\forall \pmb y'} \exp [\pmb w^T \pmb F(\pmb x, \pmb y')]} \pmb F(\pmb x, \pmb y) - \frac \lambda {|S|}  \nabla R(\pmb w)  \Big \}
\end{split}
\end{equation}
To make a comparison, we denote $\pmb s_{\pmb z}(\pmb w)$ as the (negative) SAPO update term for a sample $\pmb z$ such that
\begin{equation}\label{eq6}
\pmb w \leftarrow \pmb w -  \gamma \pmb s_{\pmb z}(\pmb w)
\end{equation}
Then, according to the procedure of SAPO algorithm, it is easy to check that $\pmb s_{\pmb z}(\pmb w)$ has the following form:
\begin{equation}\label{eq5}
\begin{split}
\pmb s_{\pmb z}(\pmb w)
&= - \Big \{ \pmb F(\pmb x, \pmb y^*) - \sum_{k=1}^{n} P_{k} \pmb F(\pmb x, \pmb y_{k}) - \frac \lambda {|S|}\nabla R(\pmb w)  \Big \}\\
&= - \Big \{ \pmb F(\pmb x, \pmb y^*) - \sum_{\forall \pmb y \in Y_n}
\frac {\exp [\pmb w^T \pmb F(\pmb x, \pmb y)]} {\sum_{\forall \pmb y' \in Y_n} \exp [\pmb w^T \pmb F(\pmb x, \pmb y')]}
\pmb F(\pmb x, \pmb y) - \frac \lambda {|S|}\nabla R(\pmb w)  \Big \}\\
\end{split}
\end{equation}
As we can see from (\ref{eq4}) and (\ref{eq5}), by increasing $n$, the SAPO update term $\pmb s_{\pmb z}(\pmb w)$ can be arbitrarily-close to the stochastic gradient $\nabla f_{\pmb z}(\pmb w)$.
More formally, we define
\begin{equation}
\delta_{\pmb z}(\pmb w)=\nabla f_{\pmb z}(\pmb w) - \pmb s_{\pmb z}(\pmb w)
\end{equation}
Then, for any $\epsilon \geq 0$, there is at least a corresponding $n$ such that,
\begin{equation}
\delta_{\pmb z}(\pmb w) \leq \epsilon
\end{equation}
In other words, when $n$ is increasing, the approximation is expected to be more and more accurate and finally reaches the point where $\delta_{\pmb z}(\pmb w) \leq \epsilon$.

\subsection{Optimum, convergence, and convergence rate}

Recall that $f(\pmb w)$ is the sequence labeling objective function, and $\pmb w \in \mathcal W$ is the weight vector. Taking the time stamp $t$ into consideration, the SAPO update (\ref{eq6}) can be reformulated as follows:
\begin{equation}\label{eq29}
\pmb w_{t+1} \leftarrow \pmb w_t - \gamma \pmb s_{\pmb z_t}(\pmb w_t)
\end{equation}
To state our convergence analysis results, we make several assumptions following \cite{Nemirovski09}. We assume $f$ is strongly convex with modulus $c$, that is, $\forall \pmb w, \pmb w' \in \mathcal W$,
\begin{equation}\label{eq30}
f(\pmb w')\geq f(\pmb w)+ (\pmb w' -\pmb w)^T \nabla f(\pmb w) + \frac c 2 ||\pmb w' - \pmb w ||^2
\end{equation}
where $||\cdot||$ means 2-norm $||\cdot||_2$ by default in this work. When $f$ is strongly convex, there is a global optimum/minimizer $\pmb w^*$.
We also assume Lipschitz continuous differentiability of $\nabla f$ with the constant $q$, that is, $\forall \pmb w, \pmb w' \in \mathcal W$,
\begin{equation}\label{eq31}
||\nabla f(\pmb w') - \nabla f(\pmb w)|| \leq q||\pmb w' - \pmb w||
\end{equation}
Also, let the norm of $\pmb s_{\pmb z}(\pmb w)$ be bounded by $\kappa \in \mathbb R^+$:
\begin{equation}\label{eq32}
||\pmb s_{\pmb z}(\pmb w) || \leq \kappa
\end{equation}
Moreover, it is reasonable to assume
\begin{equation}\label{eq33}
\gamma c < 1
\end{equation}
because even the ordinary gradient descent methods will diverge if $\gamma c >1$ \cite{conf/nips/RechtRWN11}.

Based on the assumptions, we show that SAPO converges towards the minimum $\pmb w^*$ of $f(\pmb w)$ with an arbitrary-close distance, and the convergence rate is given as follows.
%
\begin{theorem}[Optimum, convergence, and rate]\label{theo4}
	With the conditions (\ref{eq30}), (\ref{eq31}), (\ref{eq32}), (\ref{eq33}), let $\epsilon > 0$ be a target degree of convergence.
	Let $\tau$ be an approximation-based bound from $\pmb s(\pmb w)$ to $\nabla f(\pmb w)$ such that
	\begin{equation}\label{eq7}
	[\nabla f(\pmb w) - \pmb s(\pmb w)]^T (\pmb w - \pmb w^*) \leq \tau
	\end{equation}
	where $\pmb w$ is a historical weight vector that updated during SAPO training, and $\pmb s(\pmb w)$ is expected $\pmb s_{\pmb z}(\pmb w)$ over $\pmb z$ such that $\pmb s(\pmb w)= \mathbb E_{\pmb z} [\pmb s_{\pmb z}(\pmb w)]$.
	Since $\pmb s(\pmb w)$ can be arbitrary-close to $\nabla f(\pmb w)$ by increasing $n$, SAPO can use the smallest $n$ as far as the following holds:
	\begin{equation}\label{eq7.2}
	\tau \leq \frac {c\epsilon} {2q}
	\end{equation}
	Let $\gamma$ be a learning rate as
	\begin{equation}
	\gamma = \frac {c\epsilon - 2\tau q} {\beta q \kappa^2}
	\end{equation}
	where we can set $\beta$ to be any value as far as $\beta \geq 1$.
	Let $t$ be the smallest integer satisfying
	\begin{equation}\label{eq43}
	t \geq
	\frac {\beta q \kappa^2 \log {(q a_0 / \epsilon)}} {c(c \epsilon - 2\tau q)}
	\end{equation}
	where $a_0$ is the initial distance such that $a_0=||\pmb w_0 - \pmb w^*||^2$.
	Then, after $t$ updates of $\pmb w$, SAPO converges towards the optimum such that
	\begin{equation}
	\mathbb E[f(\pmb w_t)-f(\pmb w^*)]\leq \epsilon
	\end{equation}
	
\end{theorem}

The proof is in Appendix \ref{proof}.

This theorem shows that the approximation based learning like SAPO is also convergent towards the optimum of the objective function. Thus, we can approximate the true gradient by top-$n$ search and still keep the convergence properties, without having to calculate exact gradients such as the training of CRF.

More specifically, the theorem shows that SAPO is able to converge towards the optimum of the objective function with arbitrarily close distance $\epsilon$, as far as the SAPO update term $\pmb s(\pmb w)$ is a ``close-enough approximation'' (i.e., satisfying (\ref{eq7.2})) of the true gradient $\nabla f(\pmb w)$. Since $\pmb s(\pmb w)$ can be arbitrary-close to $\nabla f(\pmb w)$ by increasing $n$, SAPO can use the smallest $n$ as far as the close-enough approximation (\ref{eq7.2}) is satisfied. In practice, we find that setting $n$ to 5 or 10 already empirically satisfies the close-enough approximation in most of the real-world tasks. Moreover, the convergence rate is given in the theorem --- SAPO is guaranteed to converge with $t$ updates, and $t$ is the smallest integer satisfying (\ref{eq43}).

This analysis also explains why the structured perceptron algorithm \cite{FreundSchapire1999,Collins2002} does not converge in most of the practical tasks. As discussed before, the structured perceptron algorithm can be essentially treated as an extreme case of SAPO, which uses an extremely small $n$ as 1. In most cases, the use of $n=1$ does not satisfy the close-enough approximation condition of (\ref{eq7.2}). Thus in most cases the structured perceptron algorithm has a bad approximation over the true gradient and it diverges (as we will show in experiments).

As for the real-world tasks, the datasets are often linearly non-separable, so the structured perceptron and MIRA will diverge. However, according to the theoretical analysis, SAPO is able to remain convergent even when the data is linearly non-separable. Our experiments in Section~\ref{exp} will show how SAPO is still convergent in the real-world tasks.


\section{Experiments}\label{exp}

We describe the real-world tasks for the experiments, the experimental settings, and the experimental results as follows.

\subsection{Tasks}

We conduct experiments on natural language processing tasks with quite diverse characteristics. The natural language processing tasks include (1) part-of-speech tagging, (2) biomedical named entity recognition, and (3) phrase chunking.
All of the tasks use boolean features.
From tasks (1) to (3), the average length of samples (i.e., the number of tags per sample) is quite different, being $23.9, 26.5, 46.6$, respectively. The dimension of tags $|\mathcal Y|$ is also very diversified among tasks, with $|\mathcal Y|$ ranging from 5 to 45.

\textbf{Part-of-Speech Tagging (POS-Tag):}
Part-of-Speech (POS) tagging is an important and highly competitive task in natural language processing. We use the standard benchmark dataset in prior work \cite{Collins2002,Mochihashi2015}, which is derived from PennTreeBank corpus and uses sections 0 to 18 of the Wall Street Journal (WSJ) for training (38,219 samples), and sections 22-24 for testing (5,462 samples).
Following the previous work~\cite{Tsuruoka.ACL.09}, we use features based on unigrams and bigrams of neighboring words, and lexical patterns of the current word, with 393,741 raw features\footnote{Raw features are those observation features based only on $\pmb x$, i.e., no combination with tag information.} in total. Following previous work, the evaluation metric for this task is per-word accuracy.

\textbf{Biomedical Named Entity Recognition (Bio-NER):}
This task is from the \emph{BioNLP-2004 shared task}, which is to
recognize 5 kinds of biomedical named entities (\emph{DNA}, \emph{RNA}, etc.) on the \emph{MEDLINE} biomedical text corpus~\cite{Miwa2012,Miwa2013}. There are 17,484 training samples and 3,856 test samples.
Following the previous work~\cite{Tsuruoka.ACL.09}, we use word pattern features and POS features, with 403,192 raw features in total.
The evaluation metric is balanced F-score.

\textbf{Phrase Chunking (Chunking):}
In the phrase chunking task, the non-recursive cores of noun phrases called base
NPs are identified. The phrase chunking data is extracted from the data of
the \emph{CoNLL-2000 shallow-parsing shared task} \cite{SangBuchholz2000}. The training set consists of 8,936 sentences,
and the test set consists of 2,012 sentences. Following the previous work~\cite{Tsuruoka.ACL.09}, we use the feature templates based on word n-grams and part-of-speech n-grams, with 264,818 raw features in total.
Following previous studies, the evaluation metric for this task is balanced F-score.

\subsection{Experimental settings}

We compared the proposed SAPO algorithm with strong baselines in the existing literature, including both probabilistic gradient-based learning methods and search-based learning methods. For the probabilistic gradient-based learning methods, we choose the arguably most popular model CRF \cite{LaffertyMcCallum01}, bidirectional LSTM (Bi-LSTM)~\cite{schuster97bidirectional}, and bidirectional LSTM CRF (Bi-LSTM-CRF)~\cite{HuangEA2015} as the baselines. The CRF is with the widely used $L_2$ regularization and is trained with the standard SGD training algorithm. The Bi-LSTM and Bi-LSTM-CRF are trained with Adam optimizing algorithm~\cite{Kingma2014}.

For search-based learning methods, we choose structured perceptrons (Perc) \cite{Collins2002} and MIRA \cite{jmlr/CrammerS03}, which are arguably the most popular search-based learning methods, as the baselines.
In most cases, the averaged versions of structured perceptrons and MIRA work empirically better than naive versions of structured perceptron and MIRA \cite{Collins2002,acl/McDonaldCP05,daume06thesis,jmlr/Chiang12}. Thus we also compare SAPO with averaged versions of structured perceptrons and MIRA. To differentiate the naive and averaged versions, we denote them as Perc-Naive, Perc-Avg, MIRA-Naive, MIRA-Avg, respectively. Moreover, the MIRA method has the Nbest versions \cite{jmlr/CrammerS03,acl/McDonaldCP05}, which adopt top-$n$ search and update instead of Viterbi search and update. We also choose Nbest versions of MIRA as the additional baselines. We denote the Nbest MIRA with naive training as MIRA-Nbest-Naive and denote the one with averaged training as MIRA-Nbest-Avg.

The regularization strength $\lambda$ of CRF is tuned among values $0.1, 0.5, 1, 2, 5$ and are determined on the development data provided by the standard dataset (POS-Tag) or simply via 4-fold cross validation on the training set (Bio-NER and Chunking). With this automatic tuning for regularization strength, we set it to be $2, 5, 1$ for POS-Tag, Bio-NER, and Chunking tasks, respectively. To give no tuning advantage to SAPO, SAPO simply uses the same regularizer and the same learning rate as CRF does. All the tuning is based on CRF, and there is no additional tuning for SAPO.

Also, the proposed SAPO algorithm uses the same top-$n$ search scheme as the Nbest MIRA does.
As shown in the previous work \cite{jmlr/CrammerS03,acl/McDonaldCP05,jmlr/Chiang12}, it is good enough to use $n=5$ or $n=10$ for Nbest MIRA. It is also good enough to use $n=5$ or $n=10$ for the proposed SAPO algorithm. Thus, we set $n=5$ for Nbest MIRA and SAPO for fast speed.

The features used are based on previous work~\cite{Tsuruoka.ACL.09}. The features used for chunking are unigrams and bigrams of neighboring words, as well as unigrams, bigrams and trigrams of neighboring POS tags. The features used for Bio-NER are unigrams of neighboring chunk tags, substrings (shorter than 10 characters) of the current word, and the morphological features of the word, as well as the features used in the chunking experiments. For the features of POS-Tag, we use unigrams and bigrams of neighboring words, prefixes and suffixes of the current word, and some characteristics of the word. We also normalize the current word by turning all capital letters into lower case and converting all the numerals into `$\#$', and used the normalized word as a feature. Those features are exactly based on the previous work~\cite{Tsuruoka.ACL.09}.
%
For the neural models, the uni-gram features in the feature set use pre-trained Senna word embeddings~\cite{CollobertEA2011} with 50 dimensions for initialization. The embeddings are optimized by backpropagation during training.

The labeling scheme is the same for all models (including neural models). For POS-Tag, it is the original POS tag. For BioNER and Chunking, it is the BIO scheme following the previous work~\cite{Tsuruoka.ACL.09}.

All Experiments are performed on a computer with the Intel(R) Xeon(R) 3.0GHz CPU. For fair comparison, we set the batch size to 1 for all experiments.

\begin{figure*}[t]
	\centering
	\begin{tabular}{@{}c@{}@{}c@{}@{}c@{}@{}c@{}}
		
		\epsfig{file=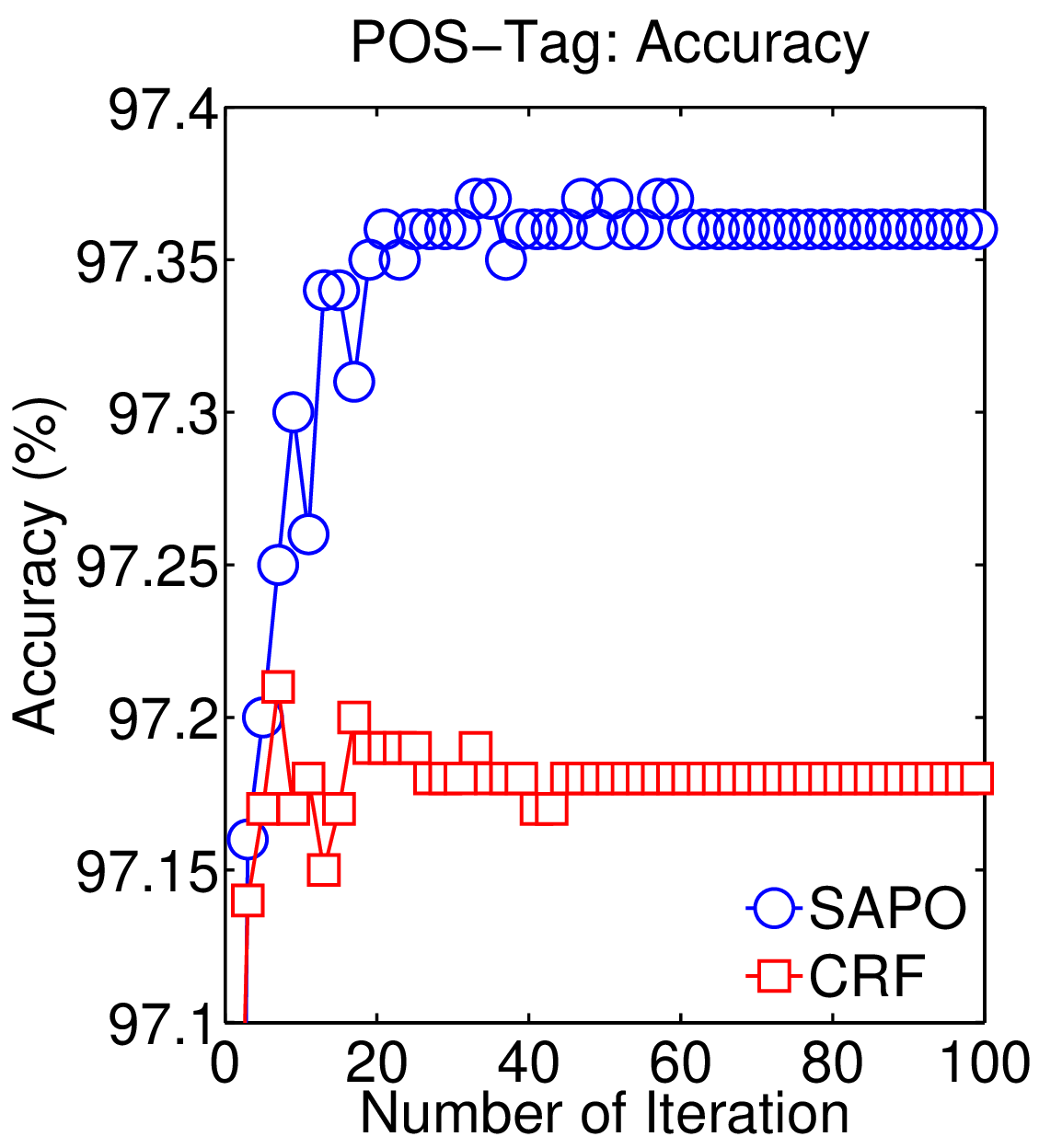,width=0.3\linewidth,clip=} &
		\epsfig{file=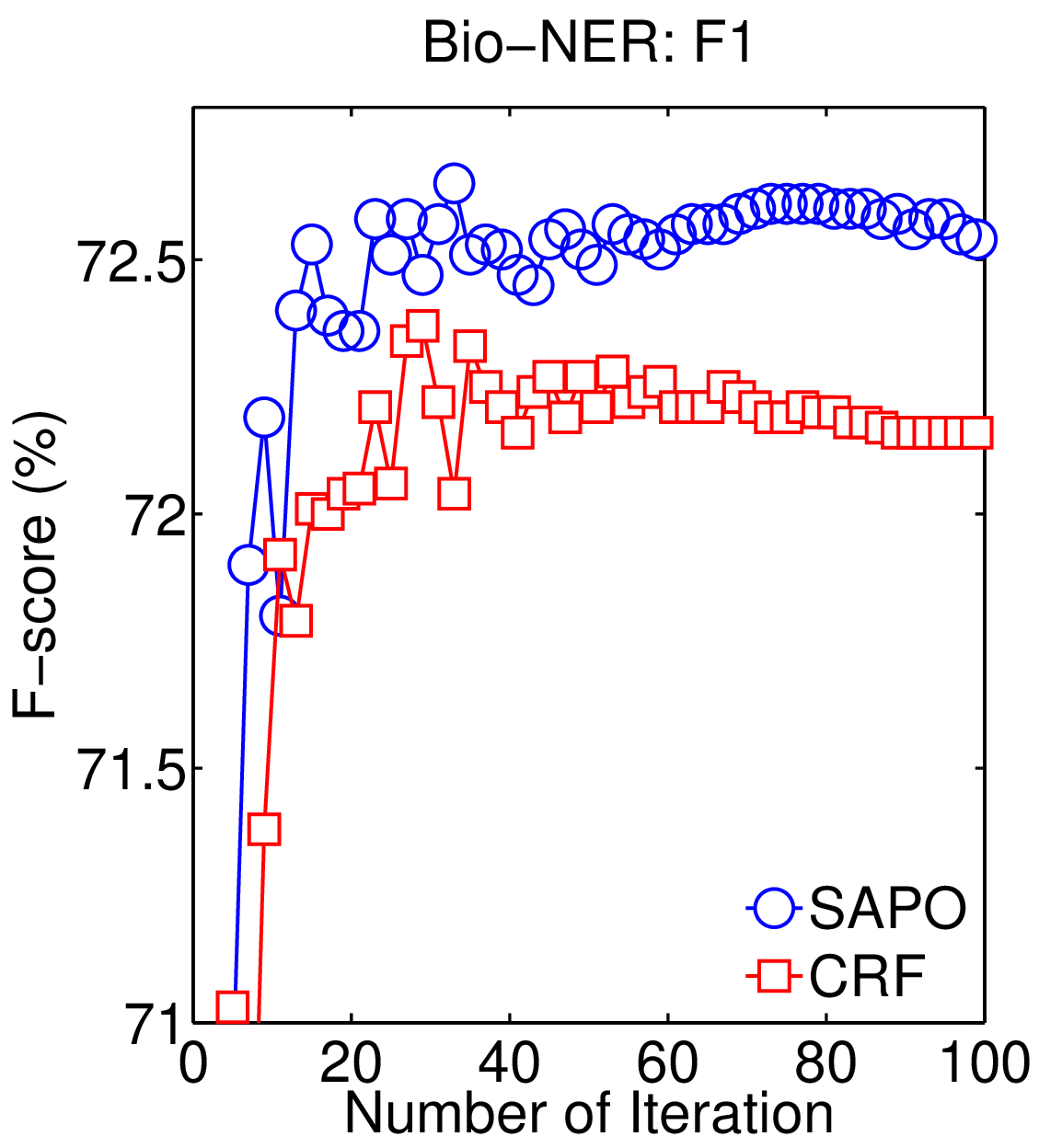,width=0.3\linewidth,clip=} &
		\epsfig{file=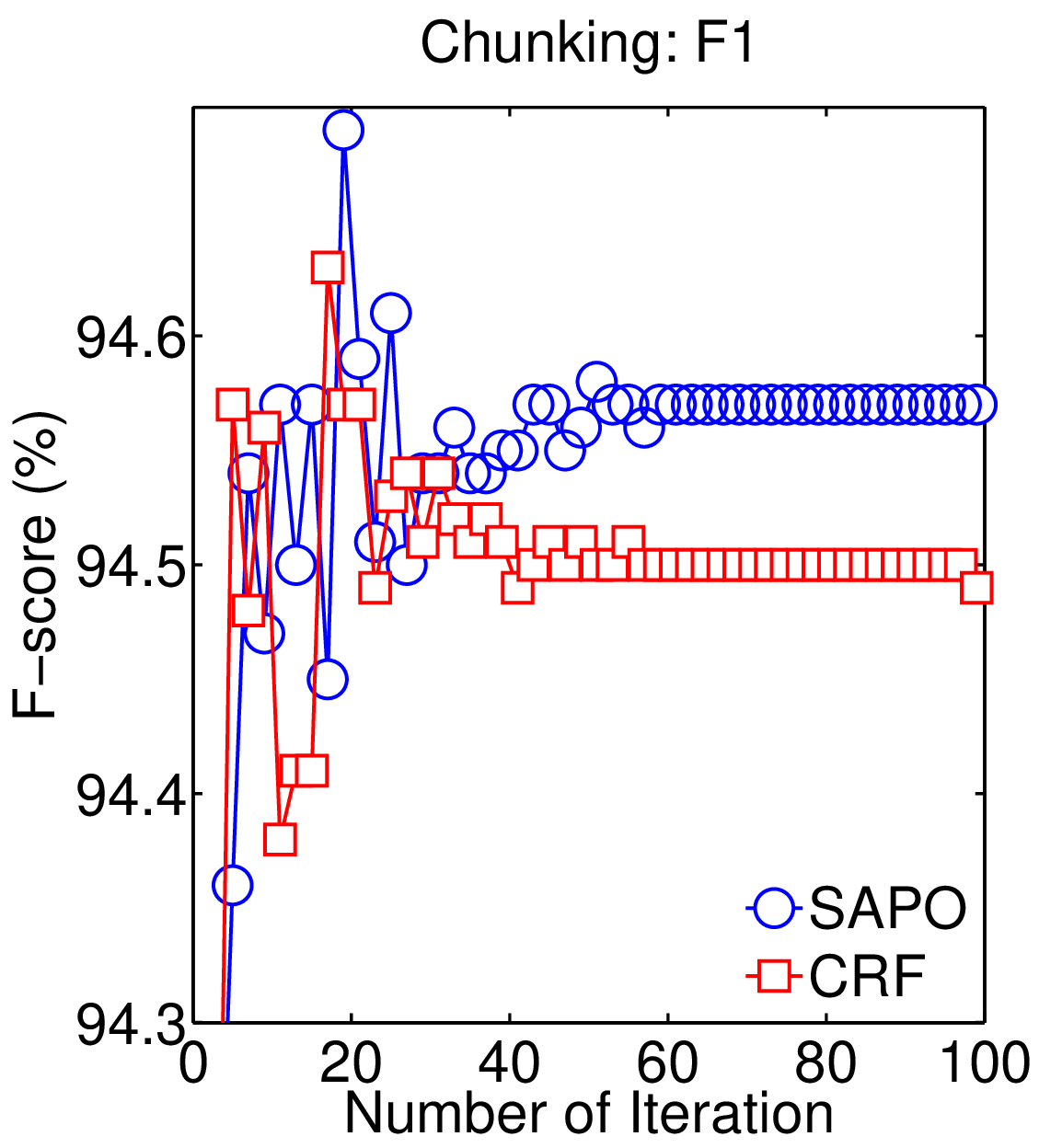,width=0.3\linewidth,clip=} \\
		
		\epsfig{file=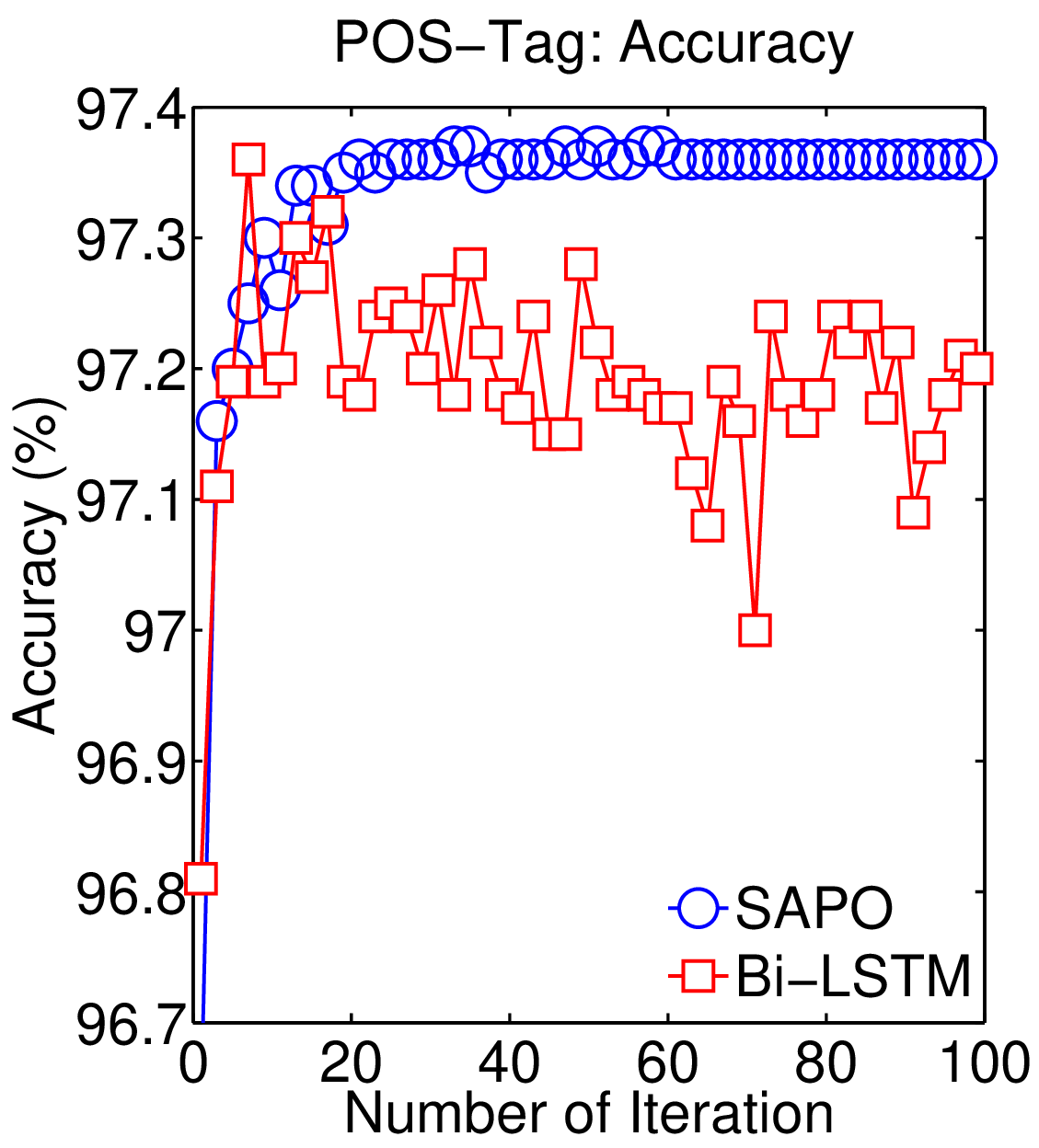,width=0.3\linewidth,clip=} &
		\epsfig{file=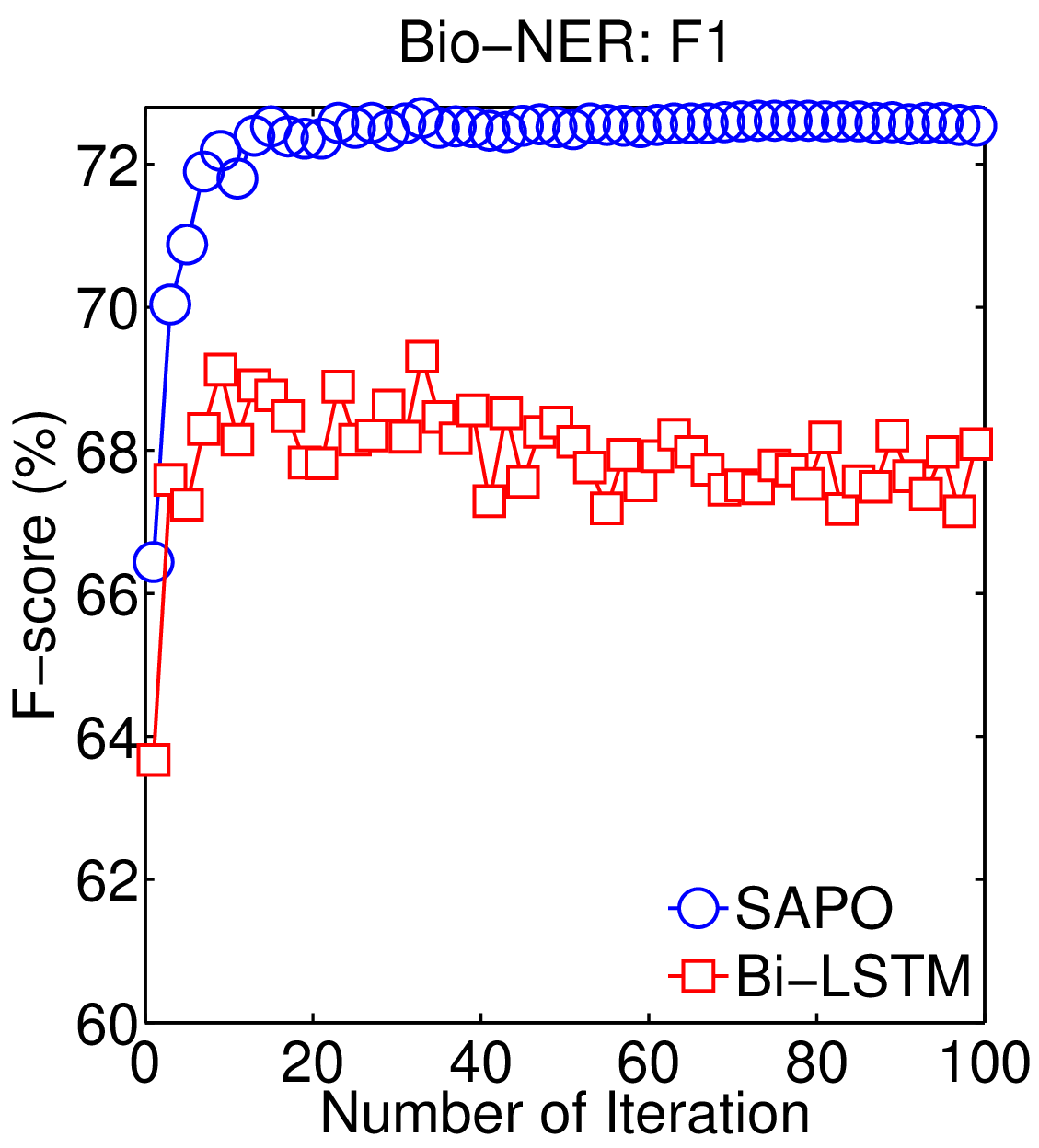,width=0.3\linewidth,clip=} &
		\epsfig{file=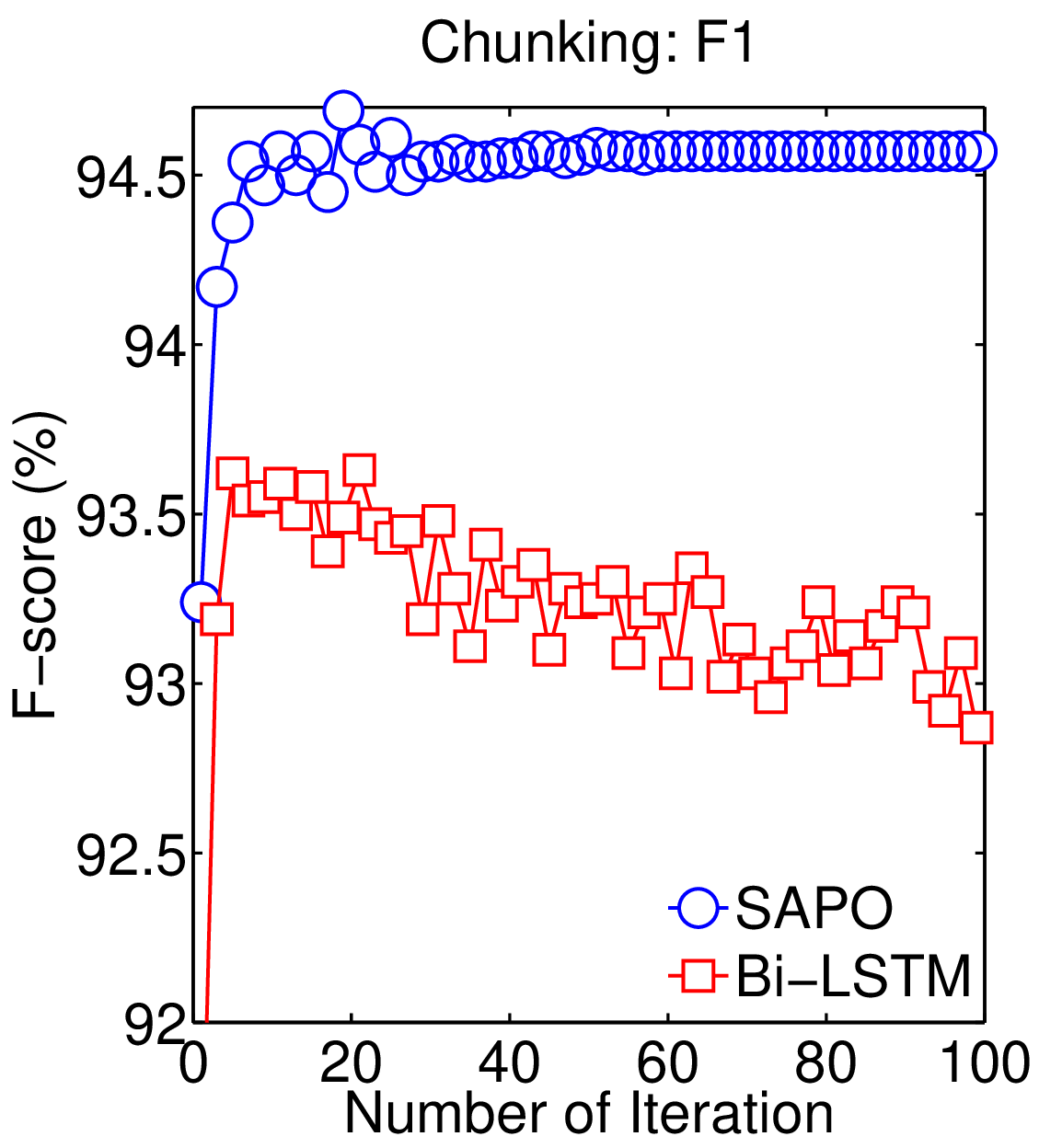,width=0.3\linewidth,clip=} \\
		
        \epsfig{file=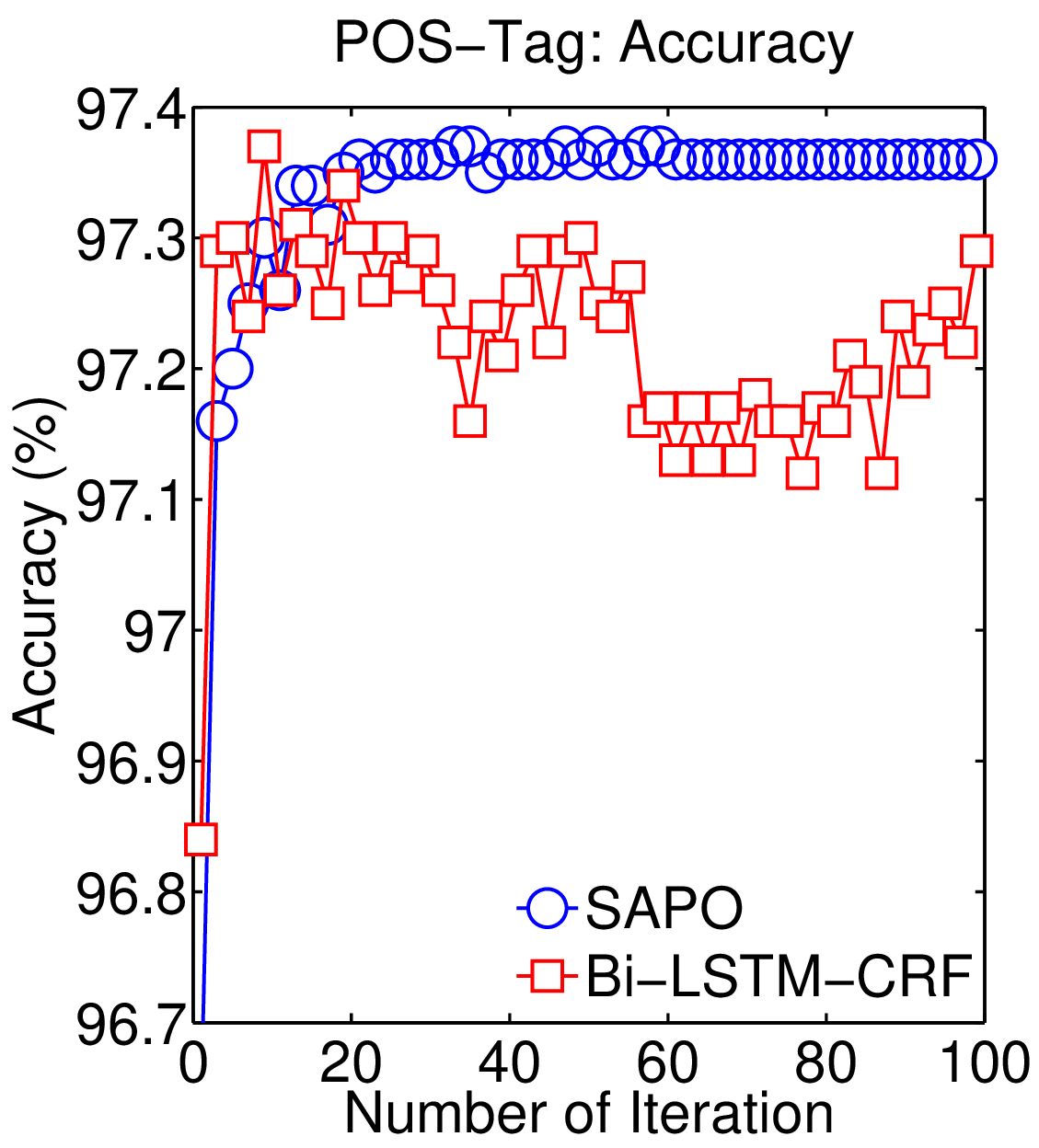,width=0.3\linewidth,clip=} &
		\epsfig{file=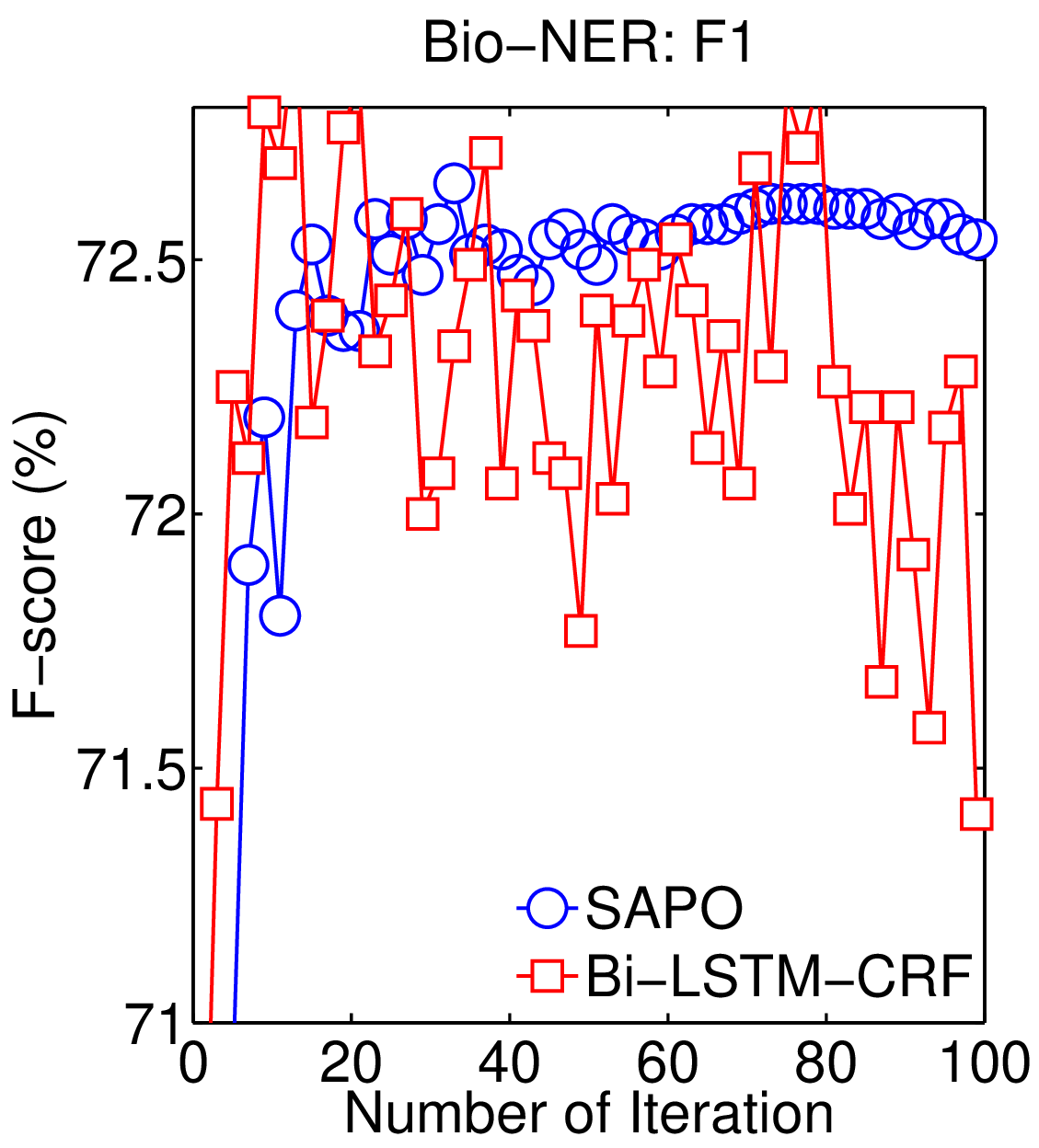,width=0.3\linewidth,clip=} &
		\epsfig{file=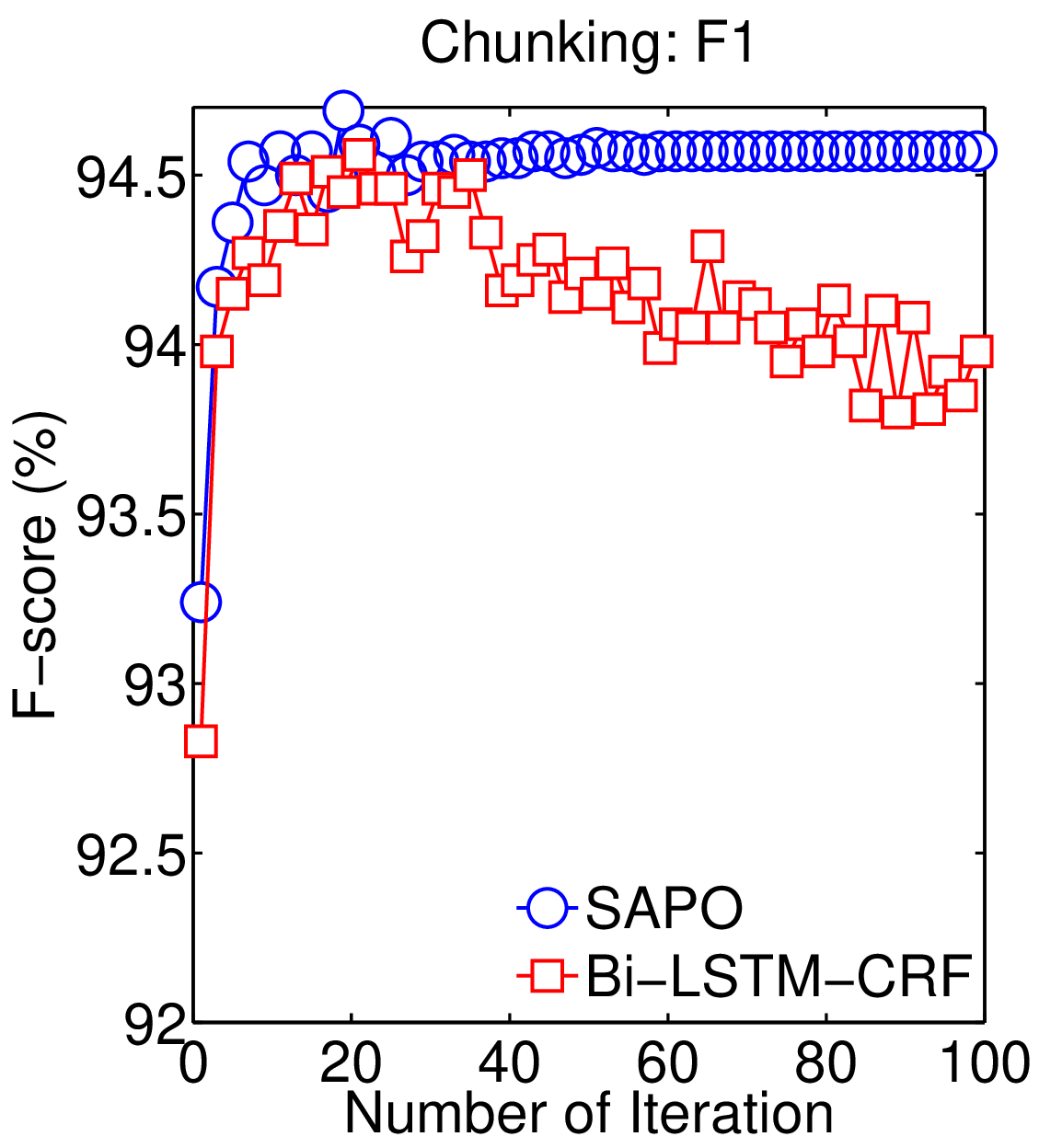,width=0.3\linewidth,clip=} \\
        
	\end{tabular}
	\caption{Comparison with CRF, Bi-LSTM, and Bi-LSTM-CRF.
	}\label{fig1}
	\vspace{-0.1in}
\end{figure*}

\begin{figure*}[t]
	\centering
	\begin{tabular}{@{}c@{}@{}c@{}@{}c@{}@{}c@{}}
		
		\epsfig{file=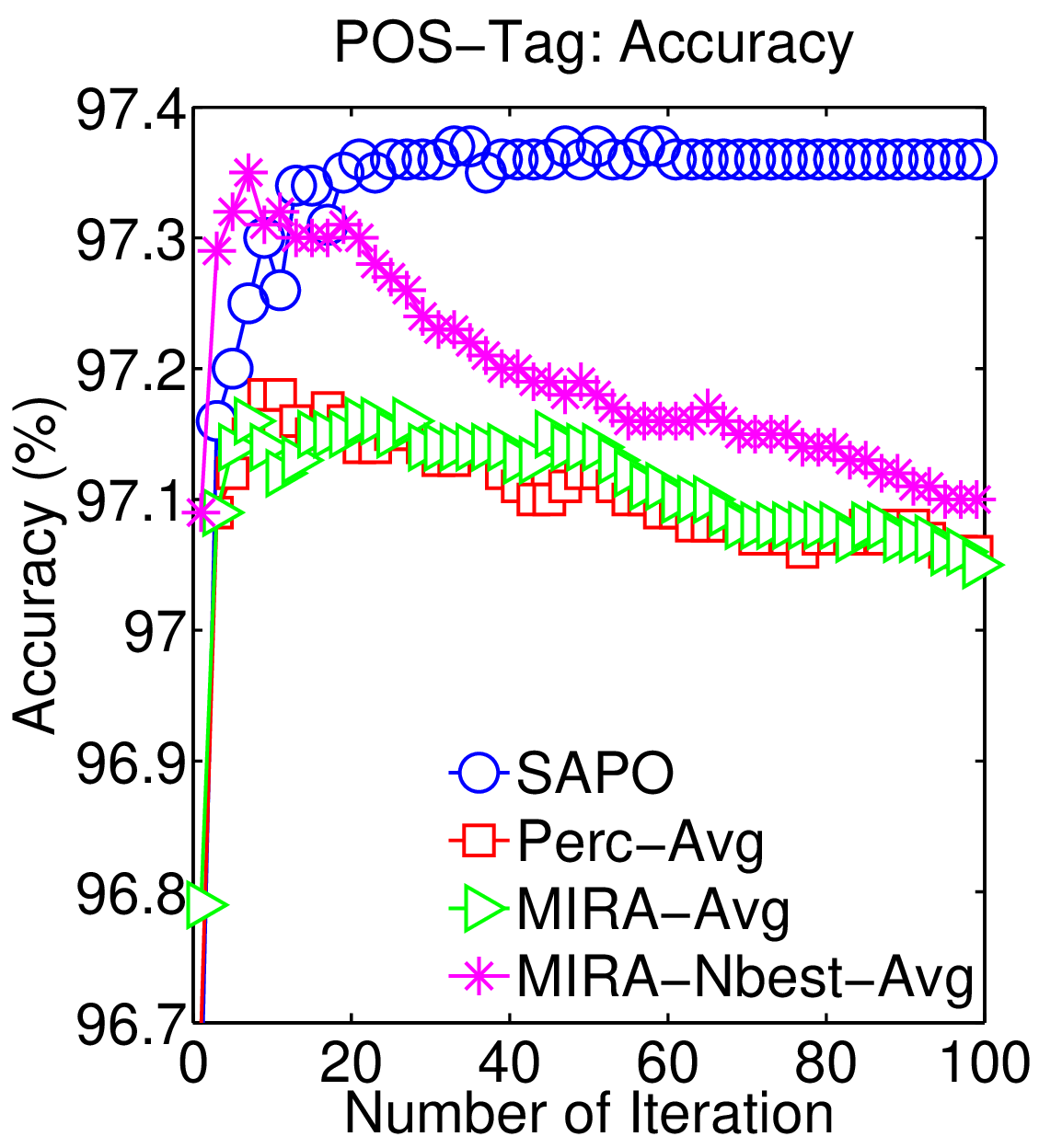,width=0.3\linewidth,clip=} &
		\epsfig{file=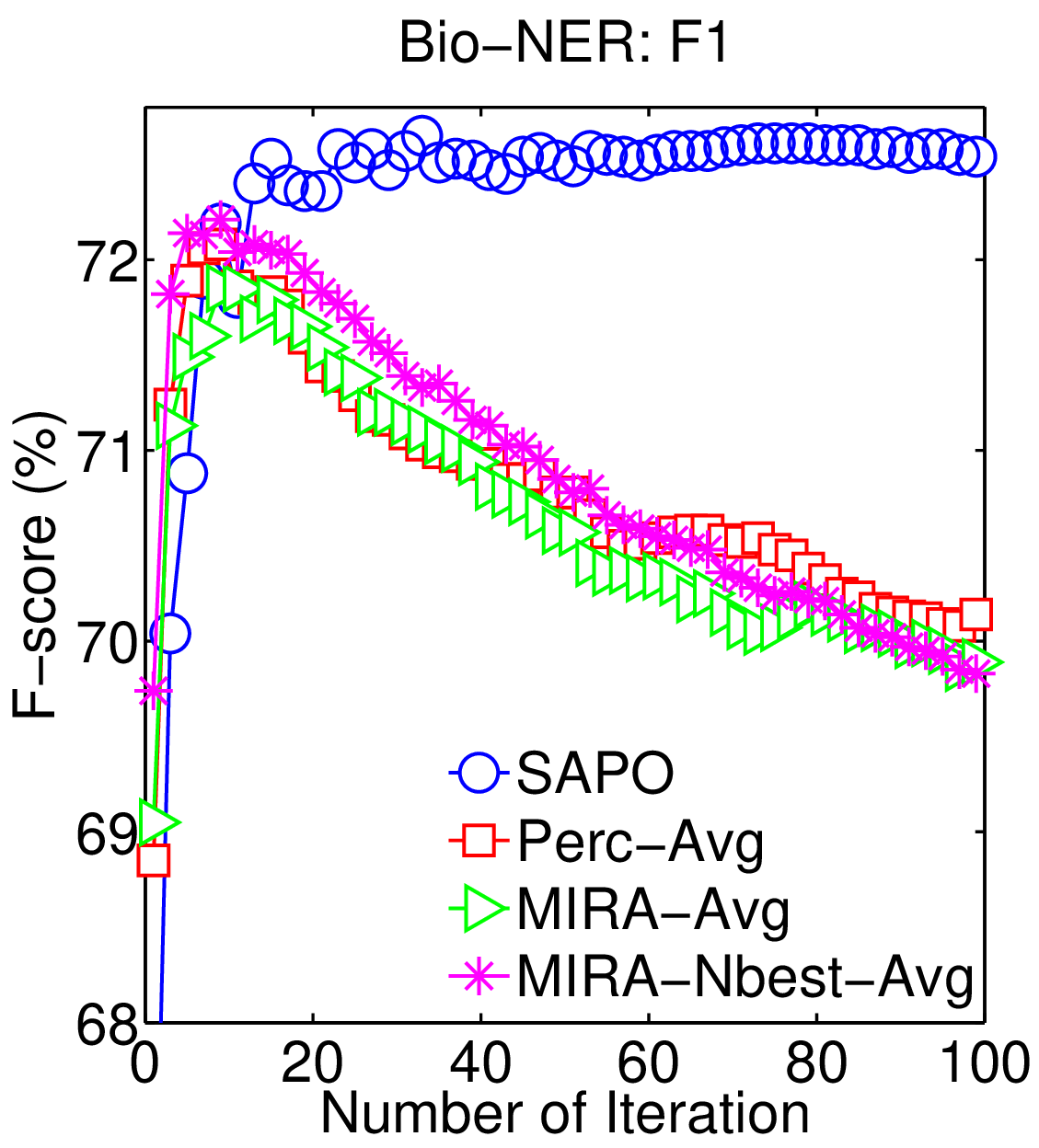,width=0.3\linewidth,clip=} &
		\epsfig{file=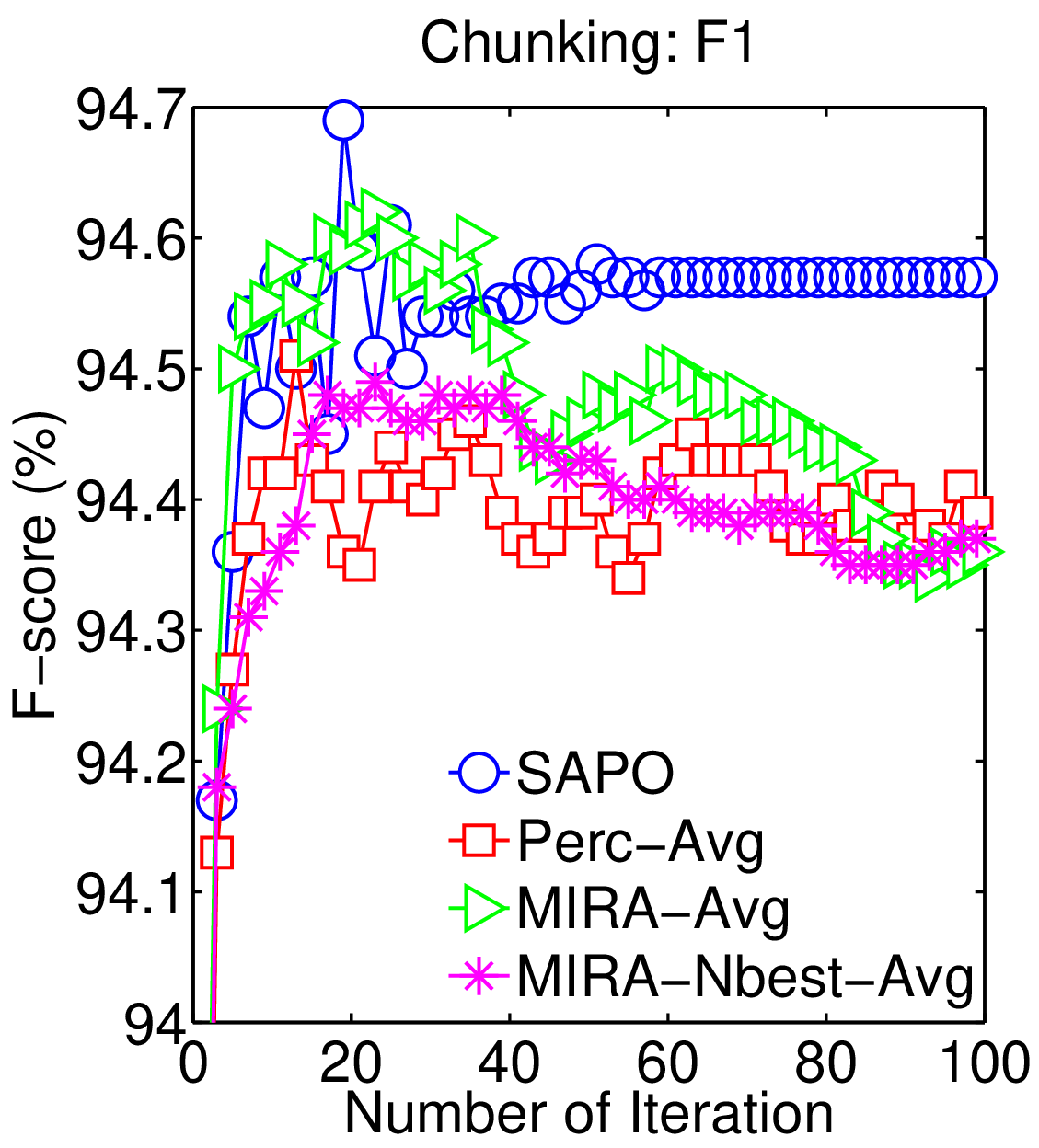,width=0.3\linewidth,clip=} \\
		
		\epsfig{file=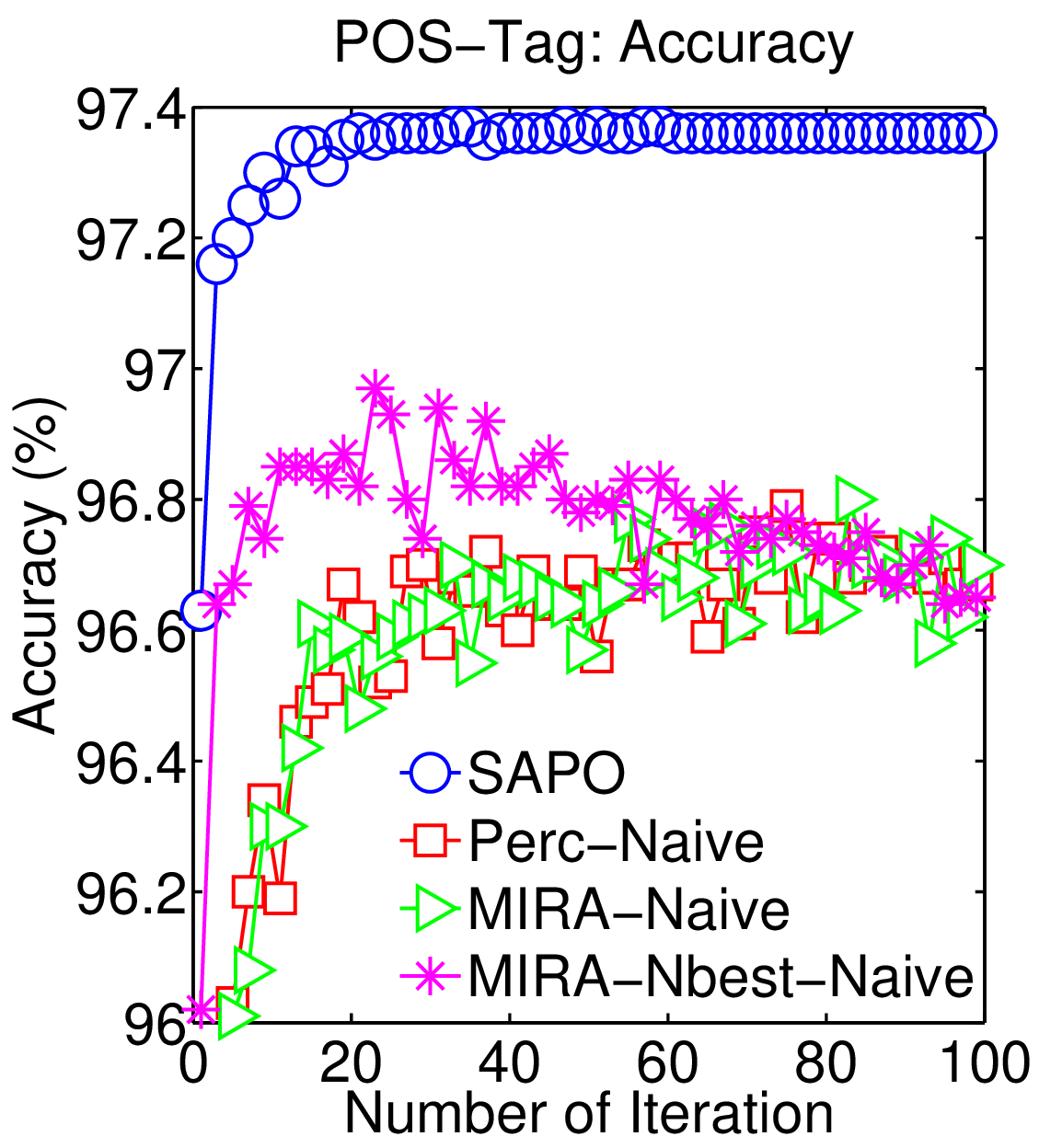,width=0.3\linewidth,clip=} &
		\epsfig{file=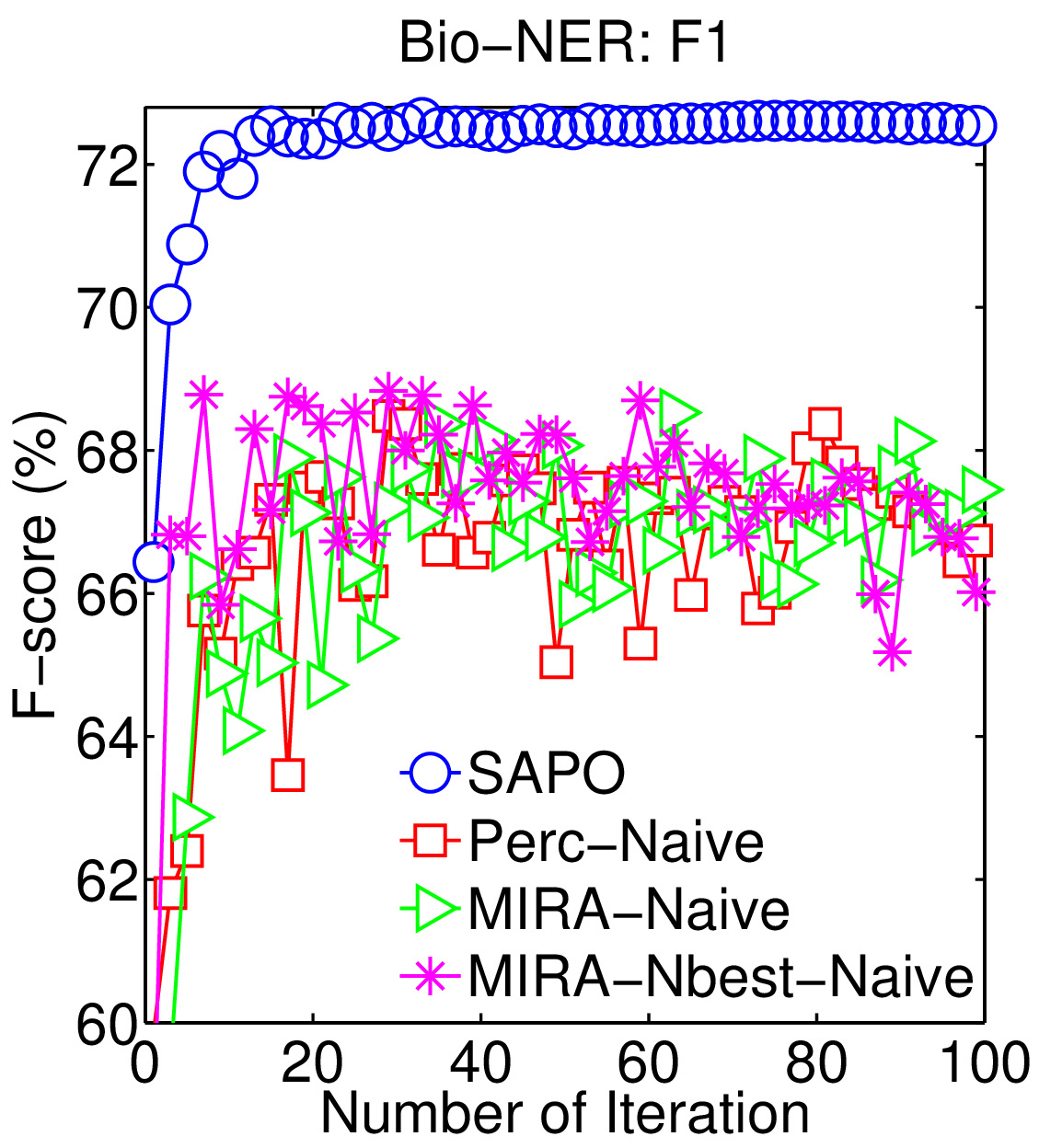,width=0.3\linewidth,clip=} &
		\epsfig{file=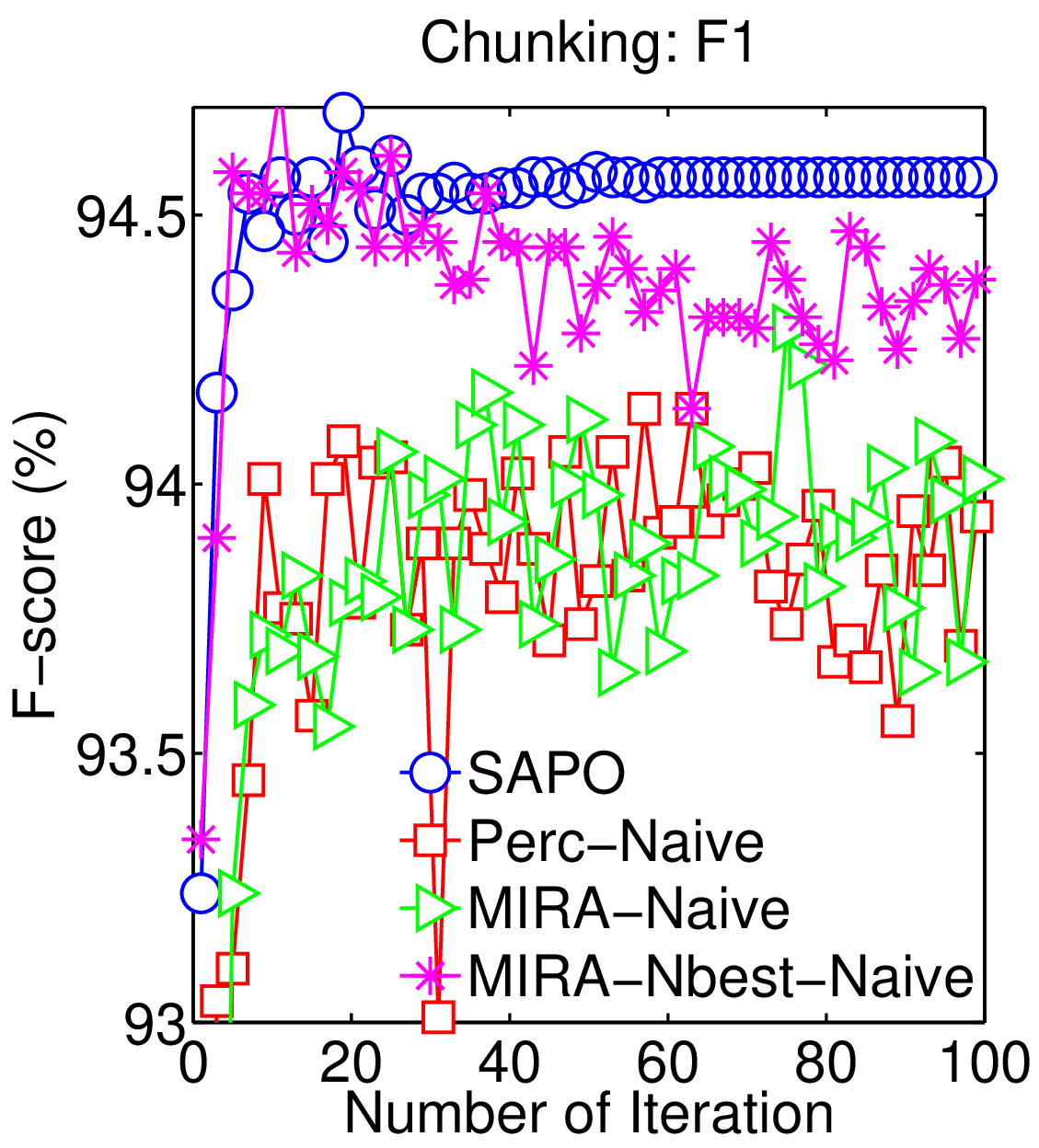,width=0.3\linewidth,clip=} \\
		
	\end{tabular}
	\caption{Comparison with structured perceptron and MIRA.
	}\label{fig2}
	\vspace{-0.1in}
\end{figure*}

\begin{figure*}[t]
	\centering
	\begin{tabular}{@{}c@{}@{}c@{}@{}c@{}@{}c@{}}

		\epsfig{file=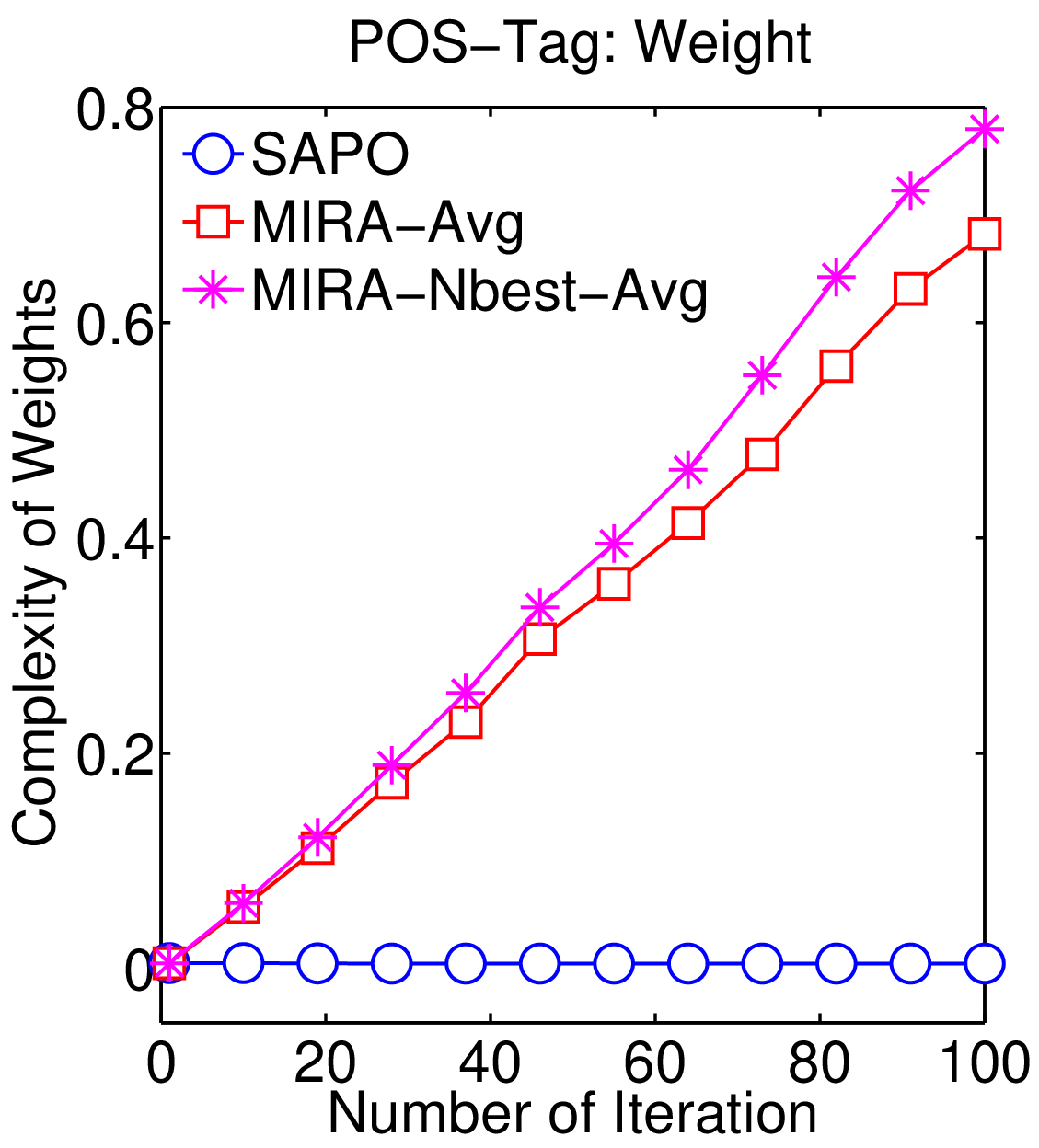,width=0.3\linewidth,clip=} &
		\epsfig{file=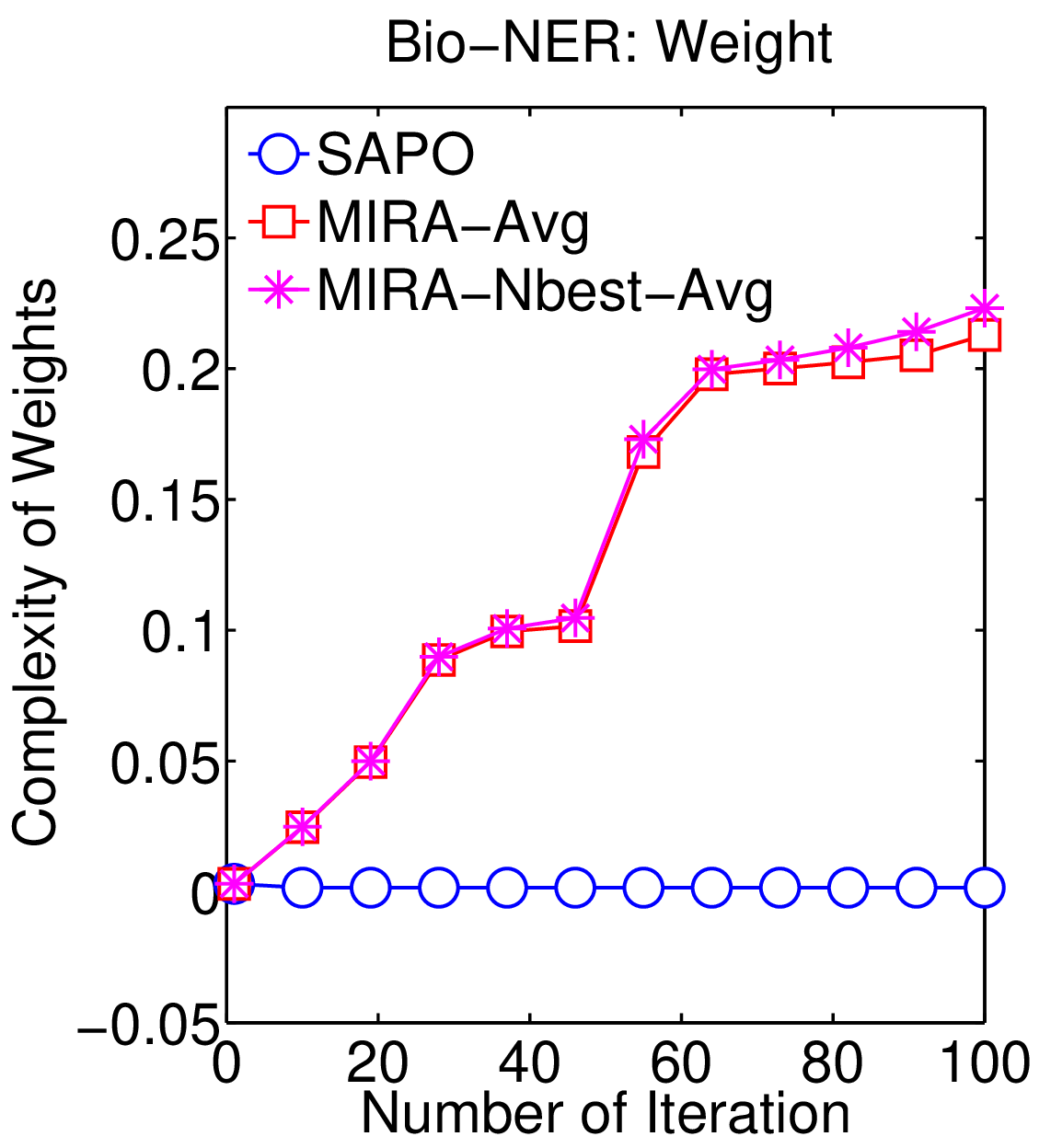,width=0.3\linewidth,clip=} &
		\epsfig{file=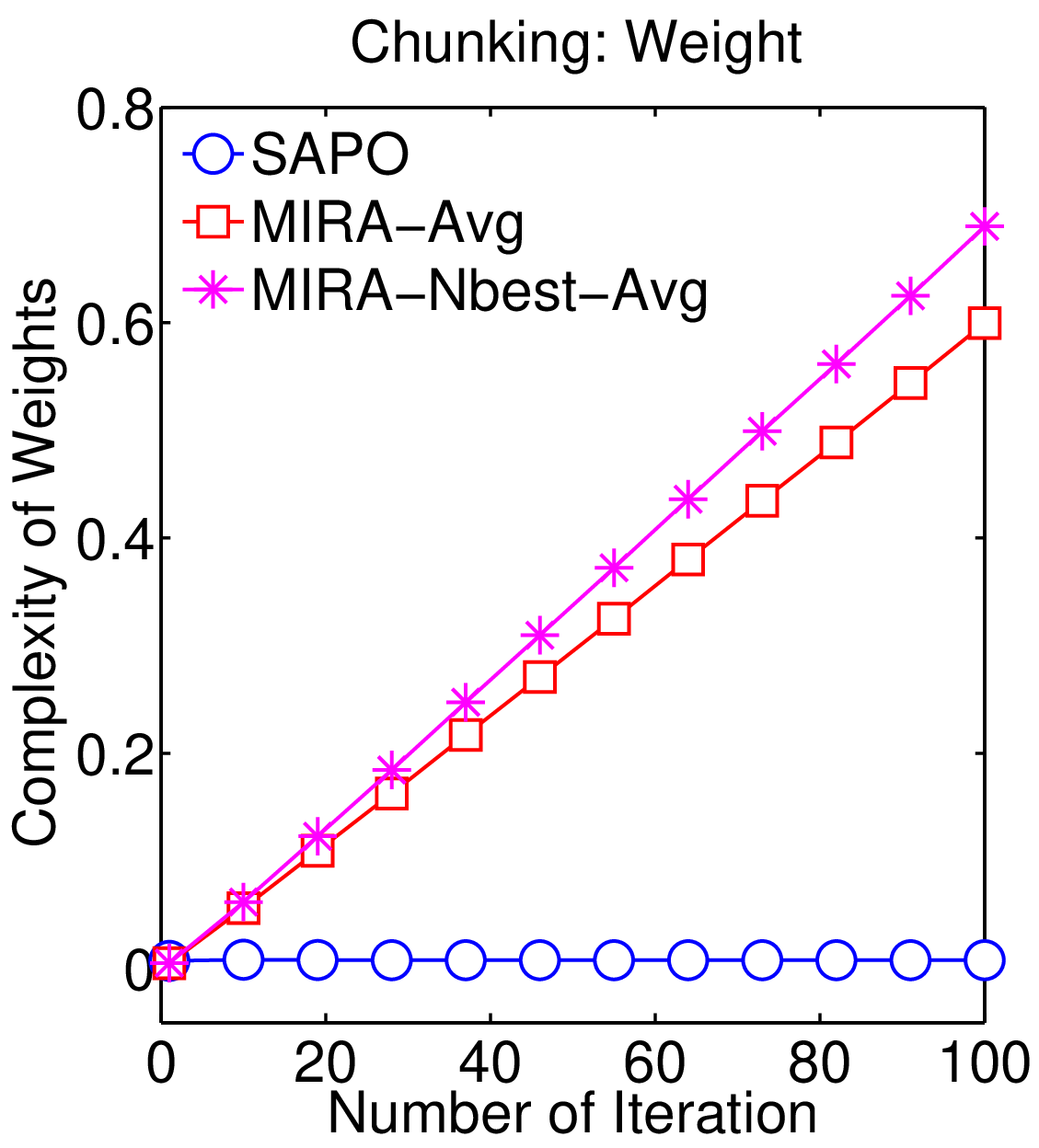,width=0.3\linewidth,clip=} \\
		
		\epsfig{file=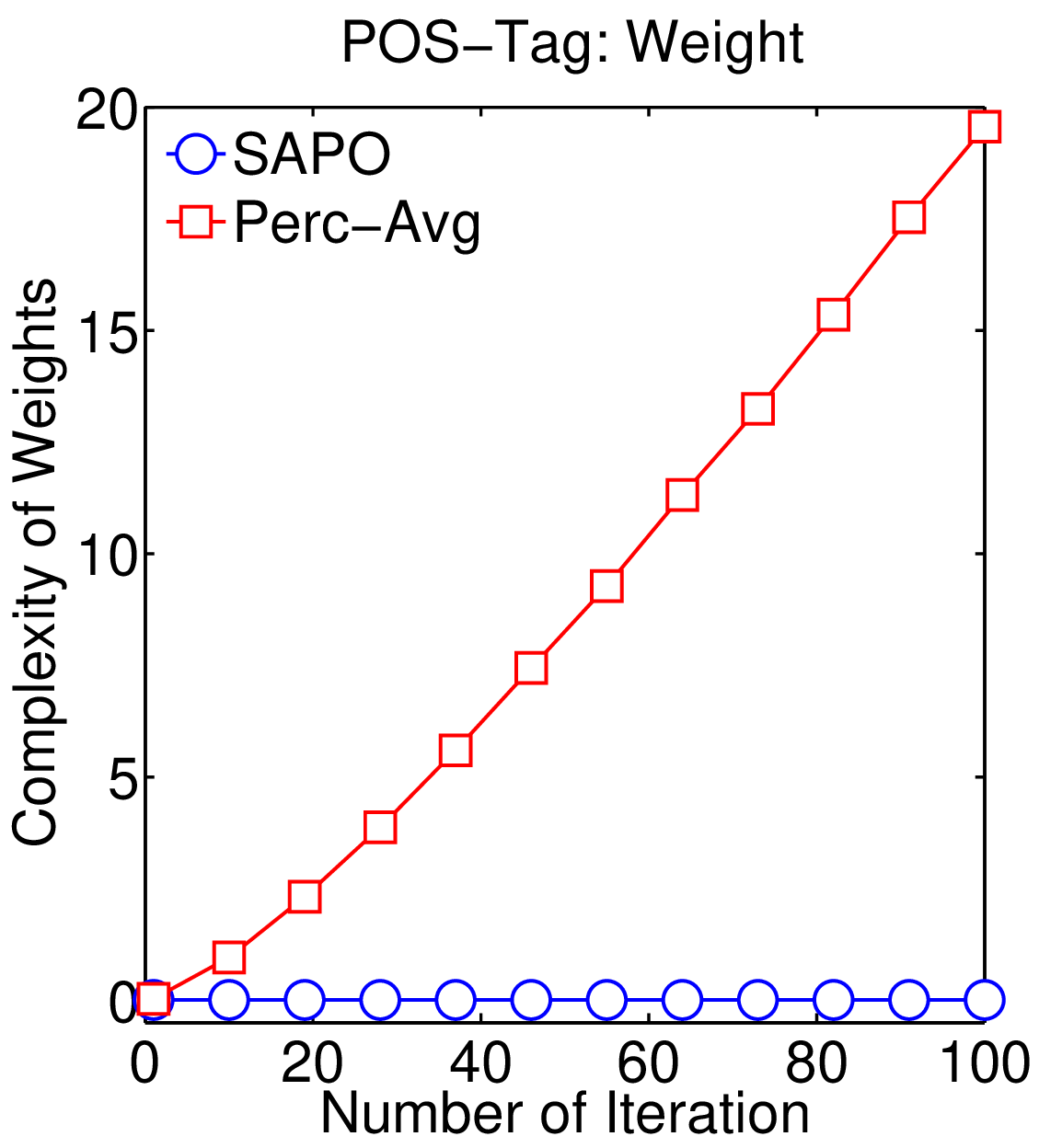,width=0.3\linewidth,clip=} &
		\epsfig{file=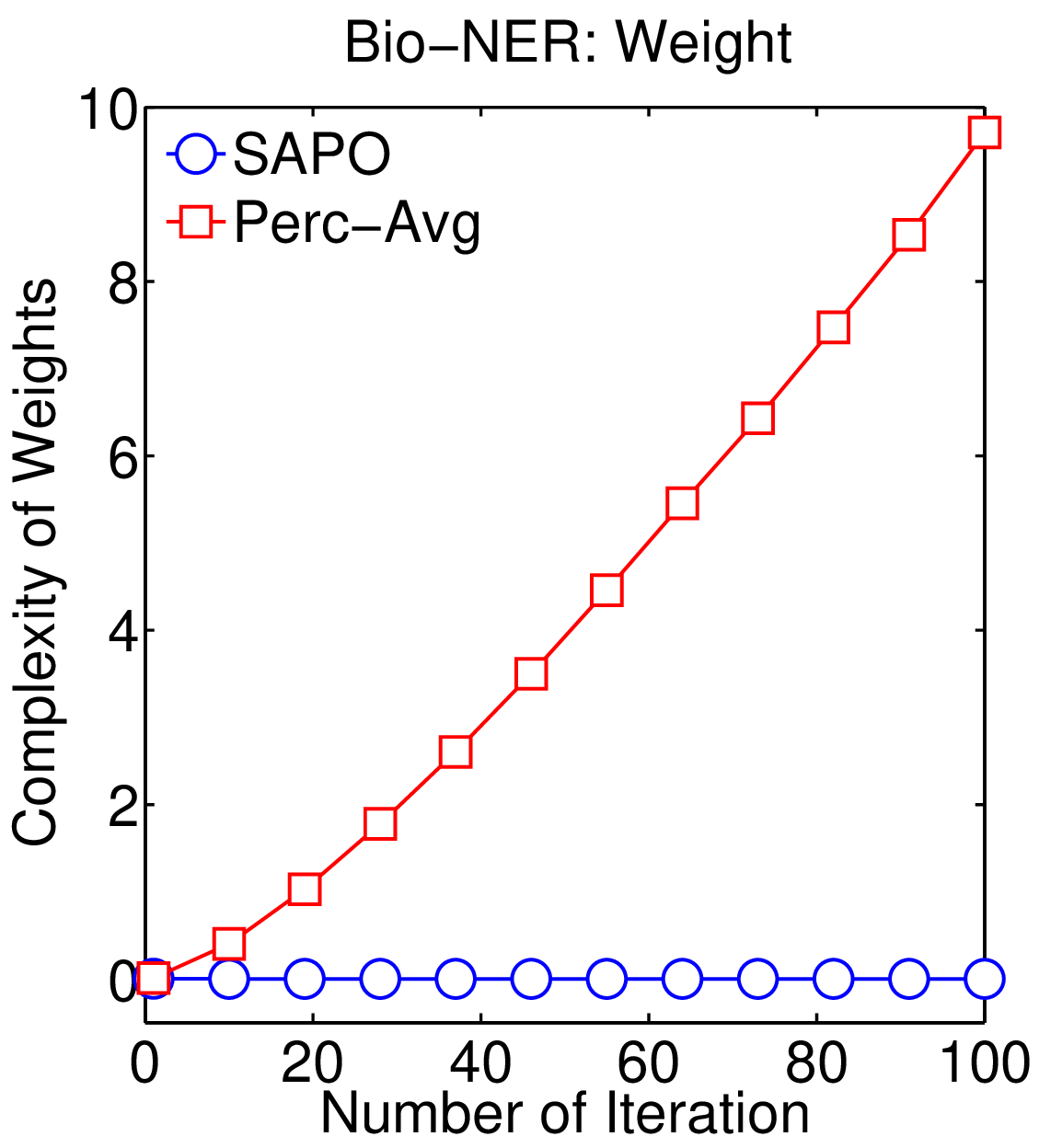,width=0.3\linewidth,clip=} &
		\epsfig{file=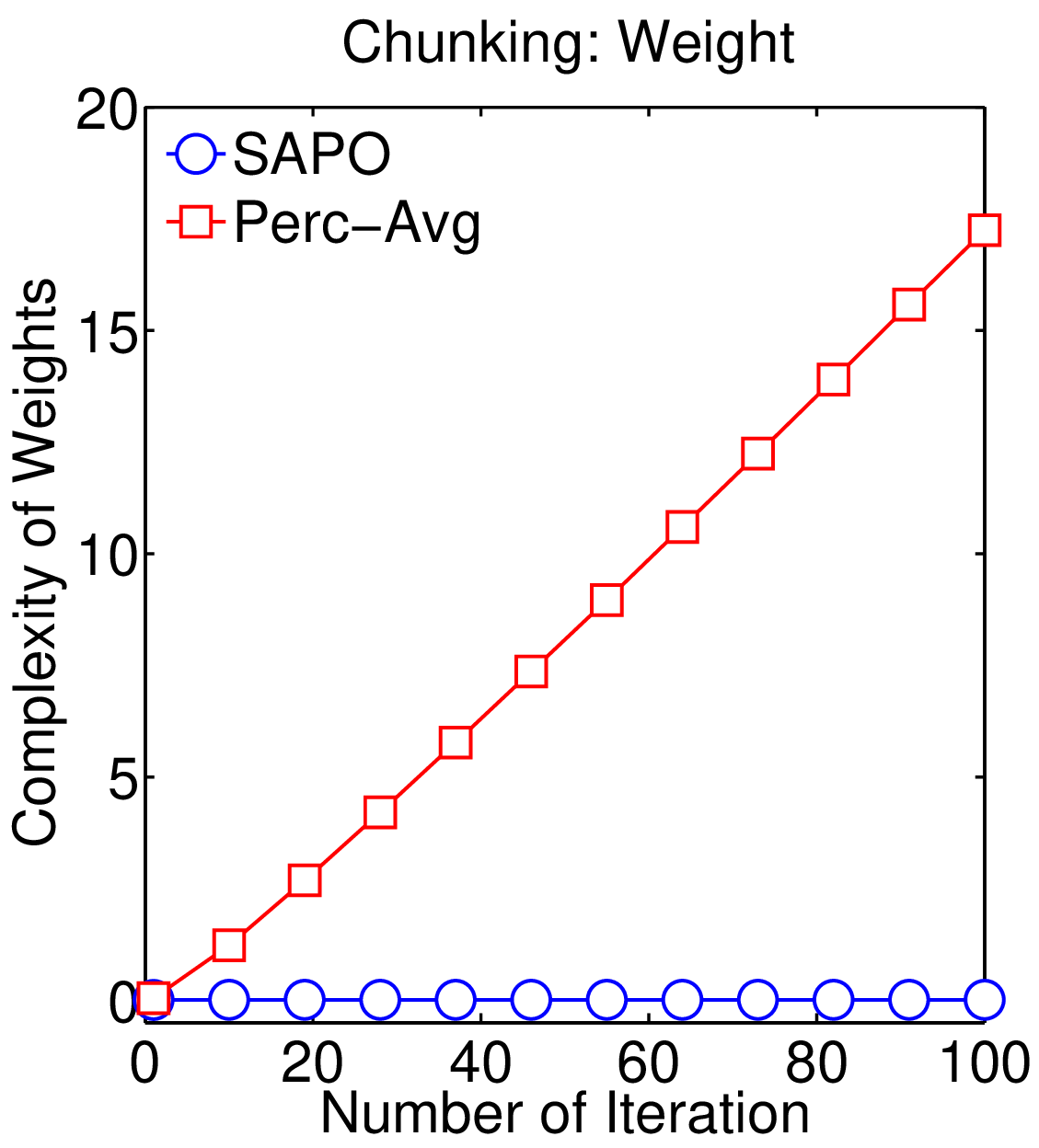,width=0.3\linewidth,clip=} \\
		
	\end{tabular}
	\caption{Comparison on parameter weights with structured perceptron and MIRA.
	}\label{fig3}
	\vspace{-0.1in}
\end{figure*}

\subsection{Experimental results}

\begin{figure*}[!h]
	\centering
	\begin{tabular}{@{}c@{}@{}c@{}@{}c@{}@{}c@{}}
		
		\epsfig{file=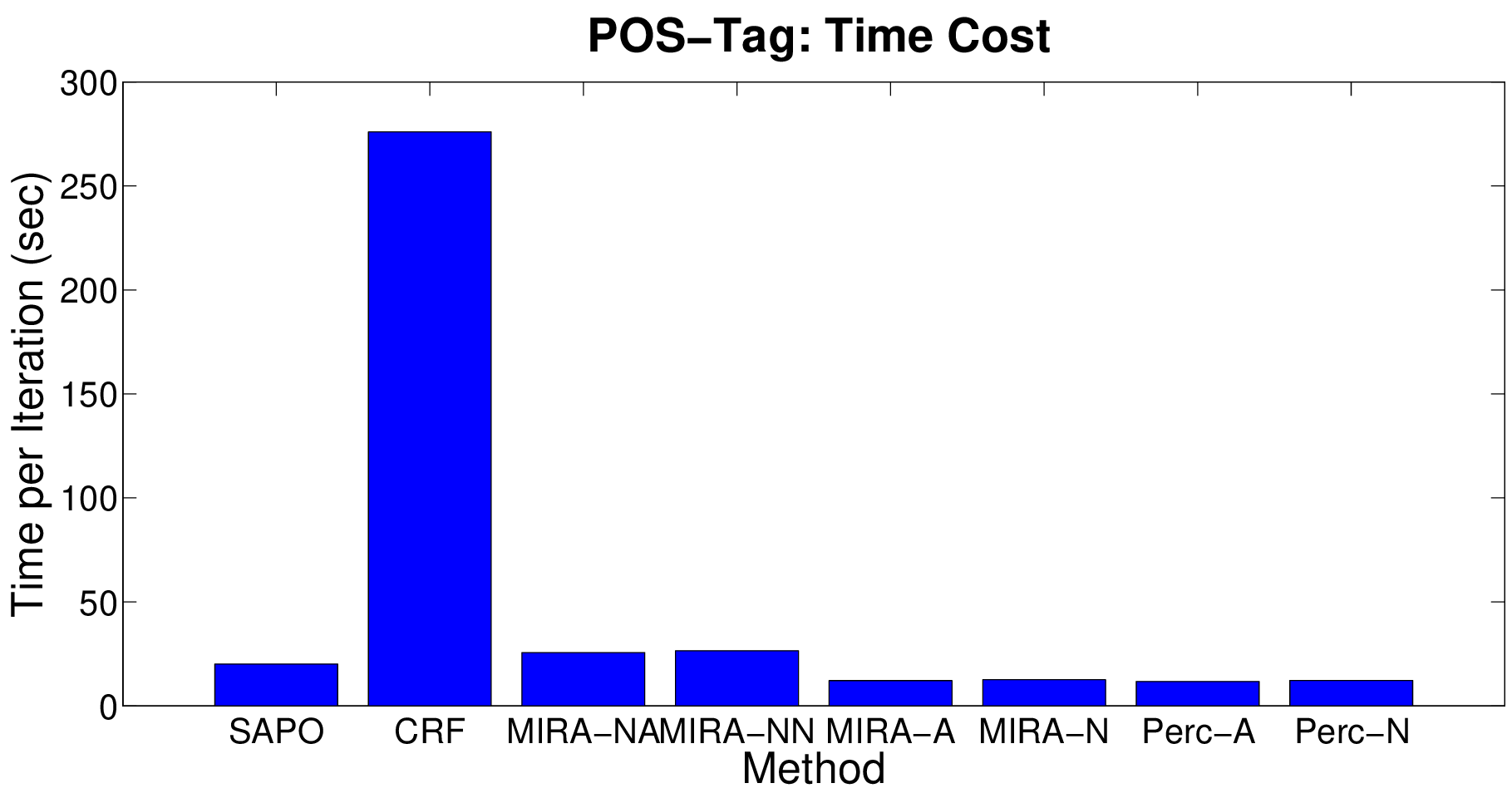,width=0.8\linewidth,clip=}\\
		\epsfig{file=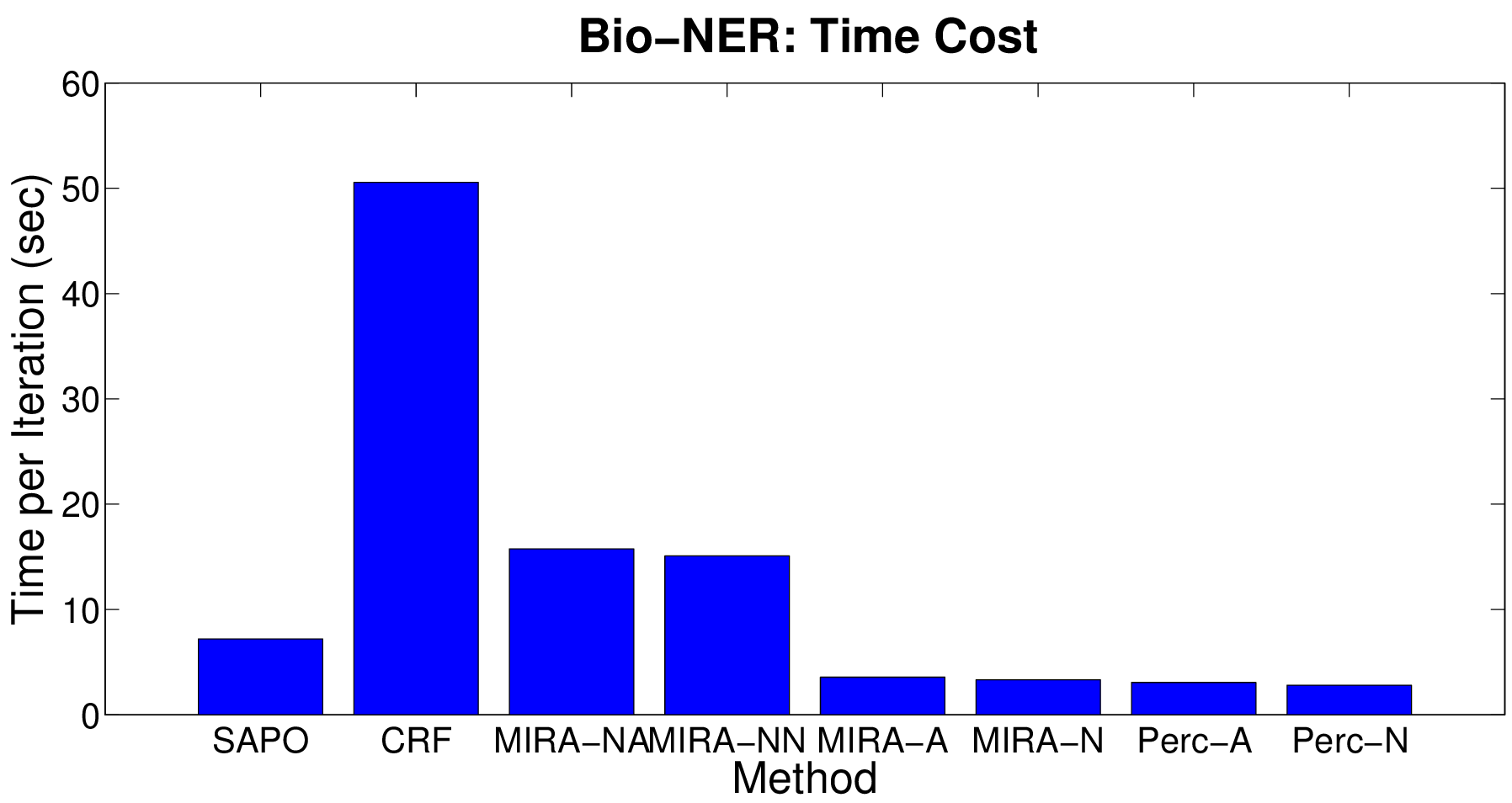,width=0.8\linewidth,clip=}\\
		\epsfig{file=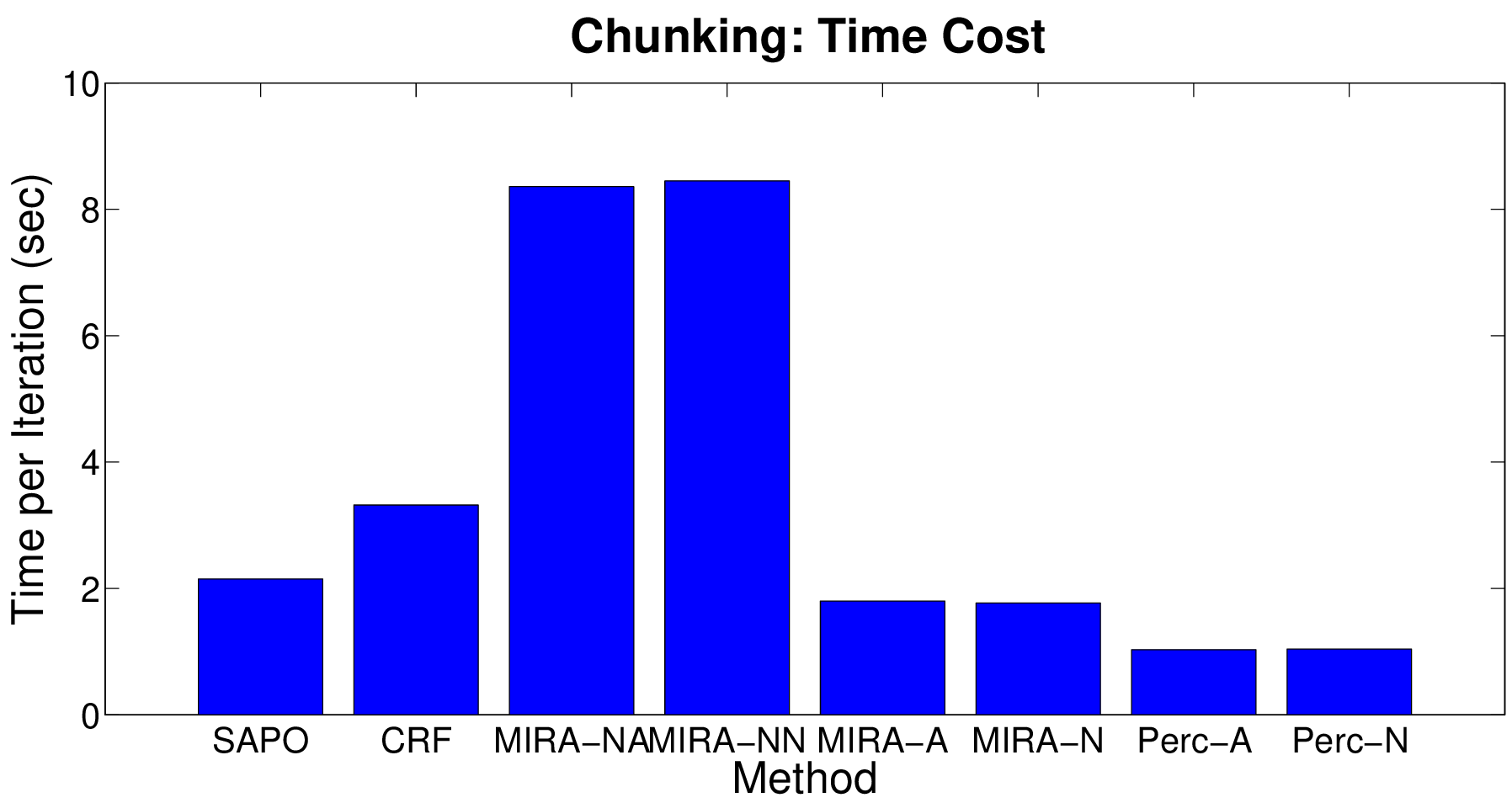,width=0.8\linewidth,clip=}\\
	\end{tabular}
	\caption{Computation time cost of different models (MIRA-NA: Average Nbest MIRA, MIRA-NN: Naive Nbest MIRA, MIRA-A: Average MIRA, MIRA-N: Naive MIRA, Perc-A: Average Perceptron, Perc-N: Naive Perceptron).
	}\label{fig4}
	\vspace{-0.1in}
\end{figure*}

The experimental results in terms of accuracy/F-score and the computation cost are shown in Figure~\ref{fig1}, Figure~\ref{fig2}, Figure~\ref{fig3}, and Figure~\ref{fig4}.
As we can see, although the tasks involve diverse feature types and different characteristics, the results are quite consistent --- the proposed SAPO algorithm has the best accuracies/F-scores in all of the three tasks compared with the existing baselines.

First, we compare SAPO with three popular gradient-based learning algorithms: CRF, Bi-LSTM, Bi-LSTM-CRF. It is impressive that the proposed SAPO algorithm even has better accuracy than the CRF, Bi-LSTM, and Bi-LSTM-CRF, which are three popular models for sequence labeling. Note that all the models are already fully optimized. As for the superiority, the reason is that the probability is distributed on top-$n$ outputs in SAPO, which is a ``regularized'' distribution instead of the probability distribution spread over all possible outputs (an exponential number). In this sense, SAPO is ``regularizing'' the exponential probability distribution to a simpler top-$n$ probability distribution. This can be seen as a probability-based regularizer with hyper-parameter $n$ controlling the regularization strength. Interestingly, the experimental results suggest that this type of regularization can indeed improve the accuracy/F-score.

We observe that SAPO is better than CRF, Bi-LSTM, and Bi-LSTM-CRF in all of the three tasks, and it shows that in many cases the differences are statistically significant.
Also, we can see that SAPO is several times faster than CRF and Bi-LSTM in terms of training time. On convergence state, SAPO performs similar or even better than CRF.

Second, we compare SAPO with search-based learning methods, including naive/average versions of Perceptron, MIRA, and Nbest MIRA. As we can see, the superiorities of SAPO over search-based learning methods are even more significant than over CRF.

We also conduct significance tests based on t-test.\footnote{For the tasks measured by F-score, the t-test is approximated by using accuracy to approximate F-score.} For the POS-Tag and chunking task, the significance test suggests that the superiorities of SAPO over all of the baselines except CRF are statistically significant, with at least $p<0.01$. For the Bio-NER task, the significance test suggests that the superiorities of SAPO over all of the baselines are significant, with at least $p<0.05$.


Figure~\ref{fig1} and Figure~\ref{fig2} show the training curves based on the number of training iterations. As we can see, SAPO and CRF are convergent as the training goes on, and Bi-LSTM, Bi-LSTM-CRF, Perc, MIRA, and Nbest MIRA diverge as the training goes on.

Figure~\ref{fig3} shows the $\pmb w$-complexity based on the number of training iterations. The $\pmb w$-complexity is the averaged (absolute) value of the weights. As we can see, SAPO is convergent and has very small weight complexity as the training goes on, and Perceptron, MIRA, and Nbest MIRA have linear or even super-linear explosion of weight complexity as the training goes on. Big weight complexity is typically a bad sign for controlling generalization risk.

\begin{figure*}[t]
	\centering
	\begin{tabular}{@{}c@{}@{}c@{}@{}c@{}@{}c@{}}
		
		\epsfig{file=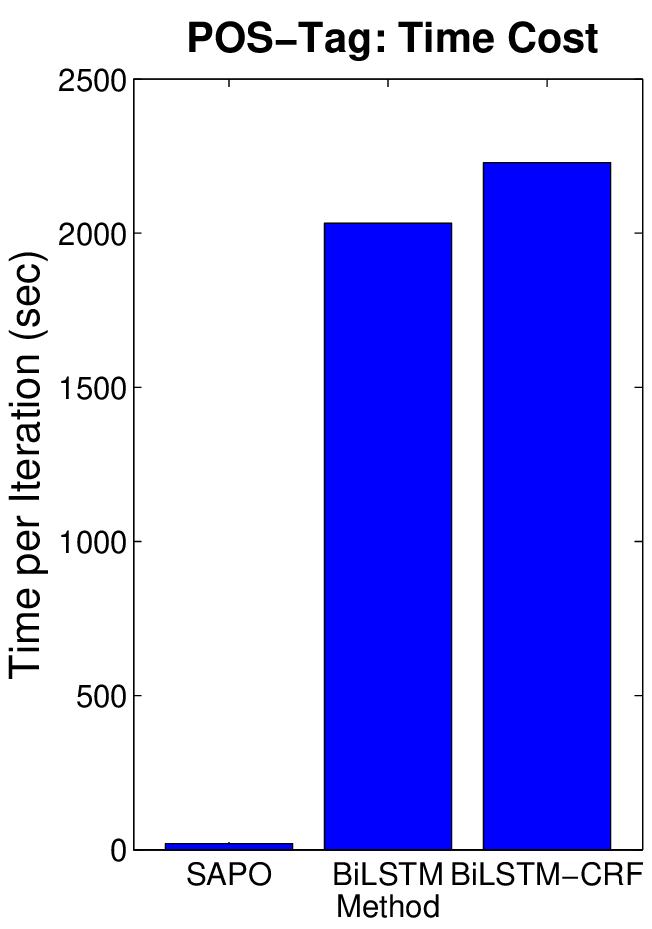,width=0.32\linewidth,clip=}&
		\epsfig{file=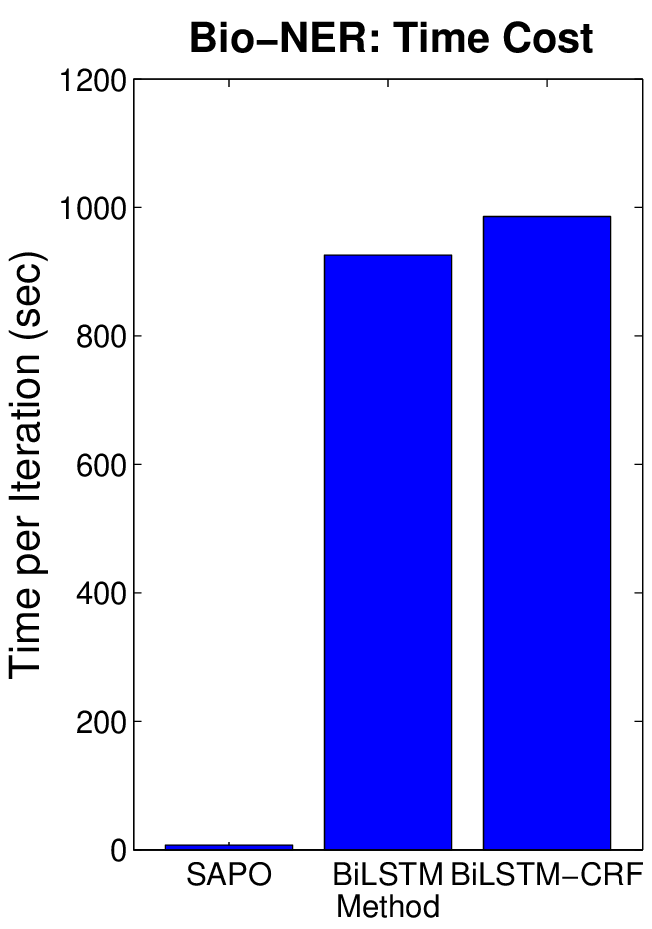,width=0.32\linewidth,clip=}&
		\epsfig{file=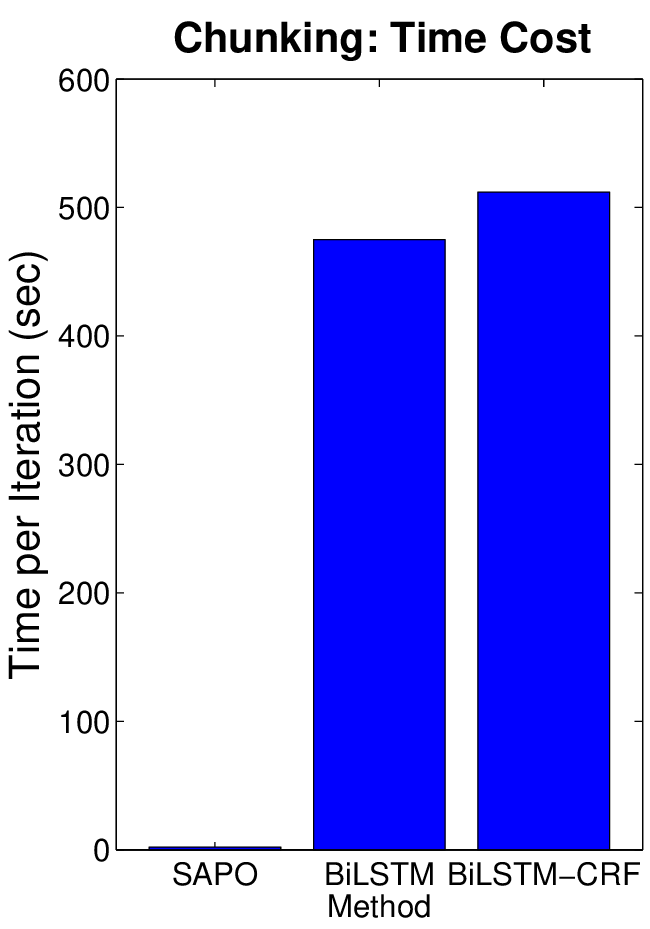,width=0.32\linewidth,clip=}\\
	\end{tabular}
	\caption{Computation time cost of SAPO, BiLSTM, and BiLSTM-CRF.
	}\label{fig5}
	\vspace{-0.1in}
\end{figure*}

Figure~\ref{fig4} and Figure~\ref{fig5} show the training time per iteration in terms of seconds. As we can see, SAPO is with low computational cost, especially compared with CRF and Bi-LSTM.

To summarize, the experiment results demonstrate that SAPO has better accuracy than probabilistic gradient-based methods like CRF and BiLSTM, at the same time with fast training speed as structured perceptrons and MIRA. 
Compared with structured perceptron, which only uses the gold sequence and causes insufficient estimation of the parameters, leading to final divergence, SAPO uses top-$n$ possible sequences with probabilistic information in training, and the parameters are more accurate.
Compared with CRF, which considers all the possible sequences but causes the overfitting problem, the $n$ in SAPO serves as a kind of regularizer and can control the complexity of the parameters.
Compared with LSTM, which is not a structured prediction model at all and does not consider the structural information, SAPO is a structured prediction model and considers the structural information.
Moreover, since LSTM can be combined with perceptrons\footnote{Most of the LSTM models for sequence labelling belong to this category.}, structured perceptrons and CRF, SAPO can also be combined with LSTM to learn the deep semantic features. 
Also, SAPO is convergent towards optimum with controllable weight complexity as the training goes on. Note that there are other important advantages of SAPO that are not revealed in those experiments --- SAPO supports search-based learning which makes gradient information not necessary and gives probability information, and it is very easy to implement.

\section{Related work}

Many sequence labeling methods have been developed including probabilistic gradient-based learning methods and search-based learning methods~\cite{wenjie2005,Mochihashi2017}. The probabilistic gradient-based learning methods include conditional random fields \cite{LaffertyMcCallum01,Tsuruoka2009}, and a variety of extensions such as dynamic conditional random fields \cite{icml04/sutton}, hidden conditional random fields \cite{QuattoniWang2008}, and latent-dynamic conditional random fields \cite{MorencyQuattoni2007}.

The search-based learning methods include margin infused relaxed algorithm \cite{jmlr/CrammerS03}, structured perceptrons \cite{Collins2002}, and a variety of related work in this direction such as latent structured perceptrons \cite{SunIJCAI09,tkde/SunML13}, confidence weighted linear classification (CW) \cite{conf/icml/DredzeCP08}, max-violation perceptrons \cite{conf/emnlp/YuHMZ13}. Most of the search-based learning methods are large-margin online learning methods.
Other related work on sequence labeling also includes maximum margin Markov networks \cite{Taskar2003} and structured support vector machines \cite{Tsochantaridis04}.

For training sequence labeling models, especially probabilistic gradient-based learning methods like CRF and its variation models, the arguably most popular training method is stochastic gradient descent (SGD) \cite{BottouC03,ZinkevichWSL10,conf/nips/RechtRWN11,conf/acl/SunWL12,SunLWL14}, which typically has faster convergence rate compared with alternative batch training methods, such as limited-memory BFGS (L-BFGS) \cite{NocedalWright1999} and other quasi-Newton optimization methods. The SGD training has theoretical guarantees to converge to the optimum weights given the convex objective function (e.g., CRF) \cite{BottouC03,ZinkevichWSL10,conf/nips/RechtRWN11}. For the search-based learning methods such as structured perceptrons, MIRA, and their variation algorithms, the training scheme is usually quite simple and self-contained in the search-based learning algorithm/model \cite{Collins2002,jmlr/CrammerS03,acl/McDonaldCP05}.

As for dealing with overfitting, the probabilistic gradient-based learning methods typically use explicit regularization terms such as the widely used $L_2$ regularizer. Other regularization schemes include the $L_1$ regularizer \cite{AndrewGao07,Tsuruoka.ACL.09}, the group Lasso regularization \cite{Yuan06mod,MartinsSFA11a}, the structure regularization \cite{Sun_NIPS2014}, and others \cite{icml/QuattoniCCD09}. For the search-based learning methods like structured perceptrons and MIRA, the scheme to deal with overfitting is less formal compared with a regularizer, usually by using parameter averaging or voting \cite{Collins2002,acl/McDonaldCP05,daume06thesis,jmlr/Chiang12}.

\section{Conclusions and future work}
The existing sequence labeling methods are problematic. The existing probabilistic gradient-based methods such as CRF and LSTM have slow training speed and do not support search-based optimization. The existing search-based learning methods such as structured perceptrons and MIRA have relatively low accuracy and are non-convergent in most of the real-world tasks. We propose a novel and ``easy" solution, a search-based probabilistic online learning framework SAPO, to address most of those issues. SAPO is with fast training, able to support search-based optimization, very easy to implement, with top accuracy, with probabilistic information, and with theoretical guarantees of convergence.

Experiments on well-known benchmark tasks demonstrate that SAPO has better accuracy than CRF and BiLSTM and roughly comparable training speed as structured perceptrons and MIRA.
Results also show that SAPO can
easily beat the strong baseline systems on those competitive tasks.

In the current implementation, our top-$n$ search uses a simple A* search algorithm with Viterbi heuristics. This top-$n$ search algorithm is not fully optimized for speed. There are several other top-$n$ search algorithms possibly with faster speed. In the future we can optimize the top-$n$ search algorithm. We believe that this can further improve the training speed of SAPO. Moreover, SAPO is a general purpose algorithm for structured classification with arbitrary structures. In the future we can apply SAPO to structured classification with structures that are more complex, e.g., syntactic parsing~\cite{wenjie2014} and statistical machine translation~\cite{zhaohai2016}.

\section*{Acknowledgements}

We thank the anonymous reviewers for their thoughtful comments. This work was supported in part by National Natural Science Foundation of China (No. 61673028).

\section*{References}

\bibliography{mybibfile}

\begin{appendices}
	\section{Appendix: Proof}\label{proof}
	
	Here we give the proof of Theorem \ref{theo4}. First, the recursion formula is derived. Then, the bounds are derived.

	\subsection{Recursion Formula}
	By subtracting $\pmb w^*$ from both sides and taking norms for (\ref{eq29}), we have
	\begin{equation}
	\begin{split}
	||\pmb w_{t+1}-\pmb w^*||^2 &= ||\pmb w_t - \gamma \pmb s_{\pmb z_t}(\pmb w_t) - \pmb w^*||^2\\
	&= ||\pmb w_t - \pmb w^*||^2 -2\gamma(\pmb w_t -\pmb w^*)^T \pmb s_{\pmb z_t}(\pmb w_t) + \gamma^2||\pmb s_{\pmb z_t}(\pmb w_t)||^2
	\end{split}
	\end{equation}
	Taking expectations and let $a_t=\mathbb E ||\pmb w_t - \pmb w^*||^2$, we have
	\begin{equation}\label{eq34}
	\begin{split}
	a_{t+1}
	&= a_t -2\gamma \mathbb E [(\pmb w_t -\pmb w^*)^T \pmb s_{\pmb z_t}(\pmb w_t)] + \gamma^2 \mathbb E [||\pmb s_{\pmb z_t}(\pmb w_t)||^2]\\
	&\text{(based on (\ref{eq32}) )}\\
	&\leq a_t -2\gamma \mathbb E [(\pmb w_t -\pmb w^*)^T \pmb s_{\pmb z_t}(\pmb w_t)] + \gamma^2 \kappa^2 \\
	&\text{(since the random draw of $\pmb z_t$ is independent of $\pmb w_t$)}\\
	&= a_t -2\gamma \mathbb E [(\pmb w_t -\pmb w^*)^T \mathbb E_{\pmb z_t}(\pmb s_{\pmb z_t}(\pmb w_t))] + \gamma^2 \kappa^2 \\
	&= a_t -2\gamma \mathbb E [(\pmb w_t -\pmb w^*)^T \pmb s(\pmb w_t)] + \gamma^2 \kappa^2 \\
	\end{split}
	\end{equation}
	We define
	\begin{equation}\label{eq8}
	\delta(\pmb w)= \nabla f(\pmb w) - \pmb s(\pmb w)
	\end{equation}
	and insert it into (\ref{eq30}), it goes to
	\begin{equation}
	\begin{split}
	f(\pmb w')
	&\geq f(\pmb w)+ (\pmb w' -\pmb w)^T [\pmb s(\pmb w) + \delta(\pmb w)] + \frac c 2 ||\pmb w' - \pmb w ||^2\\
	&= f(\pmb w)+ (\pmb w' -\pmb w)^T \pmb s(\pmb w)  + \frac c 2 ||\pmb w' - \pmb w ||^2 + (\pmb w' -\pmb w)^T \delta(\pmb w)\\
	\end{split}
	\end{equation}
	By setting $\pmb w' = \pmb w^*$, we further have
	\begin{equation}\label{eq35}
	\begin{split}
	(\pmb w - \pmb w^*)^T \pmb s(\pmb w)
	&\geq f(\pmb w) - f(\pmb w^*) + \frac c 2 ||\pmb w - \pmb w^*||^2 - (\pmb w - \pmb w^*)^T \delta(\pmb w)\\
	&\geq \frac c 2 ||\pmb w - \pmb w^*||^2 - (\pmb w - \pmb w^*)^T \delta(\pmb w)\\
	\end{split}
	\end{equation}
	Combining (\ref{eq34}) and (\ref{eq35}), we have
	\begin{equation}\label{eq36}
	\begin{split}
	a_{t+1} &\leq a_t -2\gamma \mathbb E \Big[\frac c 2 ||\pmb w_t - \pmb w^*||^2 - (\pmb w_t - \pmb w^*)^T \delta(\pmb w_t)\Big] + \gamma^2 \kappa^2 \\
	&= (1-c\gamma)a_t + 2\gamma \mathbb E [(\pmb w_t - \pmb w^*)^T \delta(\pmb w_t)] +  \gamma^2 \kappa^2
	\end{split}
	\end{equation}
	Considering (\ref{eq7}) and (\ref{eq8}), it goes to
	\begin{equation}\label{eq36.2}
	a_{t+1}
	\leq (1-c\gamma)a_t + 2\gamma \tau +  \gamma^2 \kappa^2
	\end{equation}
	We can find a steady state $a_\infty$ as follows
	\begin{equation}
	a_\infty
	=(1-c\gamma)a_\infty +  2\gamma\tau +  \gamma^2 \kappa^2\\
	\end{equation}
	which gives
	\begin{equation}\label{eq40}
	a_\infty = \frac {2\tau + \gamma \kappa^2} c
	\end{equation}
	Defining the function $A(x)=(1-c\gamma)x + 2\gamma \tau +  \gamma^2 \kappa^2$, based on (\ref{eq36.2}) we have
	\begin{equation}
	\begin{split}
	a_{t+1} &\leq A(a_t) \\
	&\text{(Taylor expansion of $A(\cdot)$ based on $a_\infty$, with $\nabla^2 A(\cdot)$ being 0)}\\
	&= A(a_\infty)+ \nabla A(a_\infty)(a_t-a_\infty)\\
	&= A(a_\infty) + (1-c\gamma)(a_t-a_\infty)\\
	&= a_\infty + (1-c\gamma)(a_t-a_\infty)
	\end{split}
	\end{equation}
	Thus, we have
	\begin{equation}\label{eq37}
	a_{t+1} - a_\infty
	\leq (1-c\gamma)(a_t-a_\infty)
	\end{equation}
	Unwrapping (\ref{eq37}) goes to
	\begin{equation}\label{eq38}
	a_t \leq (1-c\gamma)^t (a_0 - a_\infty) + a_\infty
	\end{equation}

	\subsection{Bounds}
	Since $\nabla f(\pmb w)$ is Lipschitz according to (\ref{eq31}), we have
	$$
	f(\pmb w) \leq f(\pmb w') + \nabla f(\pmb w')^T (\pmb w - \pmb w') + \frac q 2 ||\pmb w- \pmb w'||^2
	$$
	Setting $\pmb w' = \pmb w^*$, it goes to $f(\pmb w) - f(\pmb w^*) \leq \frac q 2 ||\pmb w- \pmb w^*||^2$, such that
	$$
	\mathbb E [f(\pmb w_t)- f(\pmb w^*)] \leq \frac q 2 \mathbb E ||\pmb w_t- \pmb w^*||^2 = \frac q 2 a_t
	$$
	In order to have
	\begin{equation}\label{eq10}
	E [f(\pmb w_t)- f(\pmb w^*)] \leq \epsilon
	\end{equation}
	it is required that $\frac q 2 a_t \leq \epsilon$, that is
	\begin{equation}\label{eq39}
	a_t \leq \frac {2\epsilon} q
	\end{equation}
	Combining (\ref{eq38}) and (\ref{eq39}), it is required that
	\begin{equation}
	(1-c\gamma)^t (a_0 - a_\infty) + a_\infty \leq \frac {2\epsilon} q
	\end{equation}
	To meet this requirement, it is sufficient to set the learning rate $\gamma$ such that both terms on the left side are less than  $\frac \epsilon q$. For the requirement of the second term $a_\infty \leq \frac \epsilon q$, recalling (\ref{eq40}), it goes to
	$$
	\gamma \leq \frac {c\epsilon - 2\tau q} {q \kappa^2}
	$$
	Thus, introducing a real value $\beta \geq 1$, we can set $\gamma$ as
	\begin{equation}\label{eq41}
	\gamma = \frac {c\epsilon - 2\tau q} {\beta q \kappa^2}
	\end{equation}
	Note that, to make this formula meaningful, it is required that
	$$
	c\epsilon - 2\tau q \geq 0
	$$
	Thus, it is required that
	$$
	\tau \leq \frac {c\epsilon} {2q}
	$$
	which is solved by the condition of (\ref{eq7.2}).

	On the other hand, we analyze the requirement of the first term that
	\begin{equation}
	(1-c\gamma)^t (a_0 - a_\infty) \leq \frac \epsilon q
	\end{equation}
	Since $a_0 - a_\infty \leq a_0$, it holds by requiring
	\begin{equation}
	(1-c\gamma)^t a_0 \leq \frac \epsilon q
	\end{equation}
	which goes to
	\begin{equation}\label{eq9}
	t \geq \frac {\log {\frac \epsilon {q a_0}}} {\log {(1-c\gamma)}}
	\end{equation}
	Since $\log {(1-c\gamma)} \leq -c\gamma$ given (\ref{eq33}), and that $\log {\frac \epsilon {q a_0}}$ is a negative term, we have
	$$
	\frac {\log {\frac \epsilon {q a_0}}} {\log {(1-c\gamma)}}
	\leq
	\frac {\log {\frac \epsilon {q a_0}}} {-c\gamma}
	$$
	Thus, (\ref{eq9}) holds by requiring
	\begin{equation}\label{eq42}
	\begin{split}
	t &\geq \frac {\log {\frac \epsilon {q a_0}}} {-c\gamma}\\
	&= \frac {\log {(q a_0 / \epsilon)}} {c\gamma}
	\end{split}
	\end{equation}
	Combining (\ref{eq41}) and (\ref{eq42}), it goes to
	$$
	t \geq
	\frac {\beta q \kappa^2 \log {(q a_0 / \epsilon)}} {c(c \epsilon - 2\tau q)}
	$$
	which completes the proof.
	\endproof
\end{appendices}

\end{document}